%% file: main.tex
\documentclass[manuscript,screen]{acmart}
\usepackage{algorithm}
\usepackage{algorithmic}
\usepackage{float}
\usepackage{array}
\usepackage{textcomp}
\usepackage{stfloats}
\usepackage{verbatim}
\usepackage{graphicx}
\usepackage{mdframed}
\usepackage{multirow}
\graphicspath{{Images/}}
\usepackage{blindtext}
\usepackage{subfiles} 
\usepackage{subcaption}
\usepackage{amsmath,amsfonts}

\usepackage{xcolor} 
\usepackage{tikz}
\usetikzlibrary{fit,calc}
\newcommand*{\tikzmk}[1]{\tikz[remember picture,overlay,] \node (#1) {};\ignorespaces}

\colorlet{mypink}{yellow!20!white}
\colorlet{myblue}{gray!30!white}
\usepackage{blindtext}
\usepackage{subfiles} 
\usepackage{xcolor}
\usepackage{soul}
\newif\ifhighlight

\newif\ifshowtext

\newcommand{\conditionalhighlight}[1]{%
  \ifhighlight
    \hl{#1}%
  \else
    #1%
  \fi
}

\AtBeginDocument{%
  }

\begin{document}

\title[Efficient LinearUCB for Embedded Learning Systems]{Efficient Implementation of LinearUCB through Algorithmic Improvements and Vector Computing Acceleration for Embedded Learning Systems}

\author{Marco Angioli}
\email{marco.angioli@uniroma1.it}
\orcid{0009-0002-5955-8378}

\author{Marcello Barbirotta}
\email{marcello.barbirotta@uniroma1.it}
\orcid{0000-0002-1902-7188}

\author{Abdallah Cheikh}
\email{abdallah.cheikh@uniroma1.it}
\orcid{0000-0003-4495-5960}

\author{Antonio Mastrandrea}
\email{antonio.mastrandrea@uniroma1.it}
\orcid{0000-0003-4243-1258}

\author{Francesco Menichelli}
\email{francesco.menichelli@uniroma1.it}
\orcid{0000-0002-8453-6536}

\author{Mauro Olivieri}
\email{mauro.olivieri@uniroma1.it}
\orcid{0000-0002-0214-9904}

\affiliation{%
  \institution{Department of Information Engineering, Electronics and Telecommunications (DIET),
        Sapienza University of Rome}
  \city{Rome}
  \state{Italy}
  \country{Italy}\vspace{-1em}
}

\renewcommand{\shortauthors}{Angioli et al.}

\begin{abstract}
\noindent\hrulefill

    As the Internet of Things expands, embedding Artificial Intelligence algorithms in resource-constrained devices has become increasingly important to enable real-time, autonomous decision-making without relying on centralized cloud servers.
    However, implementing and executing complex algorithms in embedded devices poses significant challenges due to limited computational power, memory, and energy resources.
    \conditionalhighlight{This paper presents algorithmic and hardware techniques to efficiently implement two LinearUCB Contextual Bandits algorithms on resource-constrained embedded devices. Algorithmic modifications based on the Sherman-Morrison-Woodbury formula streamline model complexity, while vector acceleration is harnessed to speed up matrix operations. We analyze the impact of each optimization individually and then combine them in a two-pronged strategy}. The results show notable improvements in execution time and energy consumption, demonstrating the effectiveness of combining algorithmic and hardware optimizations to enhance learning models for edge computing environments with low-power and real-time requirements.
\end{abstract}



\maketitle
\noindent\hrulefill 
    
        \subfile{Sections/1.Introduction}

        \subfile{Sections/2.Background_and_related_works}

        \subfile{Sections/3.Algorithmic_Optimizations}

        \subfile{Sections/4.Hardware_Optimizations}

        \subfile{Sections/5.Results}

        \subfile{Sections/6.Conclusions}

        \bibliographystyle{ACM-Reference-Format}
        \subfile{references.tex}

        
\end{document}

%% file: Sections/1.Introduction.tex
\section{Introduction}\label{Introduction}

    Artificial Intelligence (AI) is introducing innovative solutions and possibilities for problem-solving and automation across diverse domains, ranging from healthcare \cite{rong2020artificial} to smart manufacturing \cite{peres2020industrial}, and autonomous driving \cite{atakishiyev2021explainable}. In recent years, the rise of the Internet of Things (IoT) highlighted the necessity of embedding AI algorithms directly into small, everyday devices, transforming them into smart, autonomous units capable of reacting to the environment and making fast decisions without relying on centralized cloud servers \cite{branco2019machine,murshed2021machine}. However, implementing and running complex AI algorithms on embedded systems poses significant challenges, particularly due to their limited computational power and energy \cite{ajani2021overview}.

    \conditionalhighlight{This paper addresses these challenges, introducing algorithmic and hardware optimization strategies, which can be applied independently or combined, to reduce complexity and enhance the execution time and energy efficiency of two LinearUCB Contextual Bandits algorithms: the Disjoint and the Hybrid} \cite{li2010contextual}.
    
    On the algorithmic side, the resource-intensive process of updating and inverting the matrix underlying the learning process of these models is replaced by a more efficient, incremental update mechanism based on the Sherman-Morrison-Woodbury formula \cite{sherman1950adjustment}.
    While this principle has been adopted in previous works to implement the Disjoint LinearUCB algorithm \cite{zr_obp, contextualbandits, ciucanu2022implementing}, in this paper, we extend this technique to a more complex Hybrid variant and thoroughly assess the impact of the incremental update on the complexity, execution time, memory requirements, and energy consumption of the LinearUCB algorithms across different embedded platforms.
      
    On the other hand, looking at the hardware perspective, we leveraged the low-power Klessydra-T13 RISC-V processor's built-in vector hardware accelerator as a case study to demonstrate the impact of vector computing support on enhancing the speed of matrix operations, which are essential for optimizing algorithm performance.
    
    \conditionalhighlight{Combining these two approaches} significantly boosts the execution speed of the algorithms while reducing their complexity, memory requirements and energy consumption, representing a major advancement in their applicability in resource-constrained settings with low-power and real-time requirements.
    
    The rest of the work is organized as follows. Section \ref{Background_&_Relatedworks} presents the LinearUCB algorithms and their applications in edge computing environments.
    Section \ref{Algorithmic_optimizations} details the proposed algorithmic optimizations and discusses the achieved computational complexity reduction. 
    Section \ref{Hardware Acceleration} introduces the target RISC-V processor, focusing on the vector computing accelerator and its custom instruction set extension, which is exploited to further speed up the algorithms.
    Section \ref{Results} presents the results, discussing speedup and reduction in dynamic energy consumption. 
    Finally, Section \ref{Conclusions} summarizes the main outcomes of the work.

%% file: Sections/2.Background_and_related_works.tex
	\section{Background and Related Works}\label{Background_&_Relatedworks}

        \subsection{Background}
        Contextual multi-armed Bandits (CB) are a class of algorithms employed in sequential decision-making problems, in which the system must select the optimal action based on a given context \cite{review, li2010contextual}. These algorithms operate on line, using a learning procedure based on a rewarding mechanism. At each time step, the algorithm observes the context, picks an action and receives a reward related to the chosen option. The reward is then used to update the model parameters so that the algorithm can adapt and learn at run time, gathering enough information to predict the relationship between the observed context and the reward associated with each action. 
        \conditionalhighlight{Differently from Reinforcement learning (RL) scenarios, where actions influence subsequent states, and the objective is to maximize cumulative rewards over multiple state transitions, in CB problems, the selected action does not affect the next state, and the algorithm aims to maximize the immediate rewards at each iteration} \cite{review, angioli2023automatic, canese2021multi, littman2015reinforcement}.

        One branch of CB algorithms is the LinearUCB presented in \cite{li2010contextual}, subdivided into the Disjoint and Hybrid versions, each with different applications and advantages. The algorithms use the Upper Confidence Bound (UCB) \cite{auer2002finite} exploration strategy and assume a linear relationship between the context and the reward. At each time step $t$, the algorithm observes a set $\mathcal{A}_t$ of $N$ possible actions (or arms) together with their context feature vectors $\mathbf{x}_{t,a}$ for $a\in\mathcal{A}_t$. For a given action $a$, the expected reward is estimated based on the context $\mathbf{x}_{t,a}$, utilizing a ridge regression procedure. In the following, bold symbols are used to denote vectors and matrices, while scalar values are represented using regular (non-bold) symbols for clarity.

        The \emph{Disjoint LinearUCB algorithm} represents the simplest form of this approach, where each of the $N$ possible actions is treated as an independent entity. In this model, the action is analytically represented through a $d \times d$ matrix $\mathbf{A}_a$ and a $d \times 1$ vector $\mathbf{b}_a$, where $d$ denotes the size of the context feature vector $\mathbf{x}_{t,a}$, without shared elements among actions.
        At every step, the algorithm employs the ridge regression in (\ref{eq:ucb_disj}) to estimate the expected reward for each action and select the one with the highest value.       
        \begin{equation}\label{eq:ucb_disj}
            \begin{array}{l}
                \hat{\boldsymbol{\theta}}_{a} = \mathbf{A}_{a}^{-1} \mathbf{b}_{a}
                \\
                p_{t, a} =  \hat{\boldsymbol{\theta}}_{a}^{\top} \mathbf{x}_{t, a}+ \alpha \sqrt{\mathbf{x}_{t, a}^{\top} \mathbf{A}_{a}^{-1} \mathbf{x}_{t, a}}
            \end{array}
        \end{equation}
        Based on the chosen option, the environment yields a reward to the model, which is subsequently employed to update the matrices associated with the selected action, as outlined in (\ref{eq:update_disj}). 

        \begin{equation}\label{eq:update_disj}
             \begin{array}{l}
                \mathbf{A}_{a_{t}} = \mathbf{A}_{a_{t}}+\mathbf{x}_{t, a_{t}} \mathbf{x}_{t, a_{t}}^{\top}
                \\
                \mathbf{b}_{a_{t}} = \mathbf{b}_{a_{t}}+r_{t} \mathbf{x}_{t, a_{t}}
            \end{array}
        \end{equation}

       The \emph{Hybrid LinearUCB algorithm} is more complex than its Disjoint counterpart because it represents each action using a feature vector of size $f$, which encapsulates information common across all actions. Additionally, it employs a shared $k\times k$ matrix $\mathbf{A_0}$ and a shared $k\times 1$ vector $\mathbf{b_0}$, where $k$ is the product of the context and action feature vector lengths, $f$ and $d$. Each action in the Hybrid algorithm is modelled using a $d\times d$ matrix $\mathbf{A}_a$, a $d\times k$ matrix $\mathbf{B}_a$, and a $d\times 1$ vector $\mathbf{b}_a$. The ridge regression in this model, as outlined in (\ref{eq:ucb_hyb}), accounts for the interaction between context and action features through the $\mathbf{z_{t, a}}$ vector in $\mathbb{R}^{k}$.
        \begin{equation}\label{eq:ucb_hyb}
            \begin{array}{l}
                \hat{\boldsymbol{\beta}} = \mathbf{A}_{0}^{-1} \mathbf{b}_{0}
                \\
                p_{t, a} = \mathbf{z}_{t, a}^{\top} \hat{\boldsymbol{\beta}}+\mathbf{x}_{t, a}^{\top} \hat{\boldsymbol{\theta}}_{a}+\alpha \sqrt{s_{t, a}}
            \end{array}       
        \end{equation}
        After selecting an action, the associated and shared matrices are updated with the received reward, as per (\ref{eq:hyb_upd1}), (\ref{eq:hyb_upd2}) and (\ref{eq:hyb_upd3}).
        Equations (\ref{eq:hyb_upd1}) and (\ref{eq:hyb_upd3}) handle the updating of shared matrices, while (\ref{eq:hyb_upd2}) is specifically dedicated to updating the matrices associated with the selected action.
        \begin{equation}\label{eq:hyb_upd1}
            \begin{array}{l}
            \mathbf{A}_{0} = \mathbf{A}_{0}+\mathbf{B}_{a_{t}}^{\top} \mathbf{A}_{a_{t}}^{-1} \mathbf{B}_{a_{t}} \\
            \mathbf{b}_{0} = \mathbf{b}_{0}+\mathbf{B}_{a_{t}}^{\top} \mathbf{A}_{a_{t}}^{-1} \mathbf{b}_{a_{t}} 
            \end{array}
        \end{equation}
        
        \begin{equation}\label{eq:hyb_upd2}
            \begin{array}{l}
            \mathbf{A}_{a_{t}} = \mathbf{A}_{a_{t}}+\mathbf{x}_{t, a_{t}} \mathbf{x}_{t, a_{t}}^{\top} \\
            \mathbf{B}_{a_{t}} = \mathbf{B}_{a_{t}}+\mathbf{x}_{t, a_{t}} \mathbf{z}_{t, a_{t}}^{\top} \\
            \mathbf{b}_{a_{t}} = \mathbf{b}_{a_{t}}+r_{t} \mathbf{x}_{t, a_{t}}
            \end{array} 
        \end{equation}
        
        \begin{equation}\label{eq:hyb_upd3}
            \begin{array}{l}
            \mathbf{A}_{0} = \mathbf{A}_{0}+\mathbf{z}_{t, a_{t}} \mathbf{z}_{t, a_{t}}-\mathbf{B}_{a_{t}}^{\top} \mathbf{A}_{a_{t}}^{-1} \mathbf{B}_{a_{t}} \\
            \mathbf{b}_{0} = \mathbf{b}_{0}+r_{t} \mathbf{z}_{t, a_{t}}-\mathbf{B}_{a_{t}} \mathbf{A}_{a_{t}}^{-1} \mathbf{b}_{a_{t}}
            \end{array}
        \end{equation}

        The Hybrid algorithm imposes considerably higher memory requirements and computational demands, potentially presenting challenges for its direct deployment in resource-constrained embedded environments. However, including shared matrices in the Hybrid model enables the estimation of rewards for context-action tuples that have not been tested yet but are similar to those previously explored \cite{li2010contextual}. This enhancement substantially accelerates the convergence time for the algorithm, particularly in scenarios with strong correlations among actions.

        \conditionalhighlight{Notably, unlike RL algorithms such as Q-learning } \cite{review,canese2021multi}\conditionalhighlight{, which update state-action values with scalar adjustments, LinearUCB algorithms heavily rely on matrix operations at each step, motivating the optimizations introduced in this work.}

        \subsection{Related Works}
        Contextual Bandits algorithms have demonstrated their versatility across a wide range of applications. 
        In healthcare and biology, these algorithms have been applied to improve treatment allocation in clinical trials \cite{pmlr-v85-durand18a} and propose personalized dosing strategies \cite{MD2}.         
        In recommendation systems, they found notable applications in personalizing article selection or tailoring movie recommendations like for Yahoo homepages \cite{li2010contextual} and Netflix \cite{Netflix} respectively, demonstrating the ability to learn user preferences over time, enhancing the experience and engagement \cite{he2020contextual, li2010contextual, review, tewari17ads}.
        Furthermore, recent advancements have shown the potential of LinearUCB algorithms in enhancing hardware efficiency. The study in \cite{angioli2023automatic} employs these algorithms to automatically select the optimal configuration of a general-purpose hardware accelerator according to the workload and reconfigure the architecture at run-time. 

        \conditionalhighlight{Recently, Contextual Bandit algorithms have also emerged in edge computing environments for localized data processing and real-time decision-making.
        The work in }\cite{tewari17ads} \conditionalhighlight{proposes using CB algorithms directly executed on mobile phones or wearable devices for mobile health, personalizing interventions based on user context such as GPS location, calendar busyness, and heart rate. This approach ensures interventions can occur even in network issues or failures, providing robust and timely support. It also addresses privacy concerns by keeping sensitive health data localized on the user's device.
        In} \cite{dronebandit} \conditionalhighlight{, authors employ CB algorithms on resource-constrained nanodrones to dynamically allocate inference tasks between edge and cloud, based on predicted network delays. This approach demonstrates improved adaptability in task allocation, enabling efficient computation offloading and optimized edge computing in fluctuating network conditions.
        The work in }\cite{chen2018spatiotemporal}\conditionalhighlight{ proposes a CB approach to optimize the placement of services on edge nodes, enhancing response times and service quality for users. Their algorithm runs directly on edge devices and dynamically rents computing resources based on spatial and temporal demand patterns, adapting to changes in user needs across different times and places.}
        \cite{yang2021edge} \conditionalhighlight{ proposes a CB algorithm to split tasks across multiple edge devices, selecting the most reliable ones in real-time. This ensures stable, deadline-driven performance even in fluctuating network conditions, making it ideal for large-scale, distributed edge applications. }
        In the context of emerging networks, authors in \cite{6GTechnology} utilize CB to optimize intelligent and secure radio environments in 6G vehicular-aided heterogeneous networks, demonstrating their potential to enhance the efficiency and security of these advanced networks. In all these contexts. \conditionalhighlight{All these applications require CB algorithms to be implemented efficiently and meet the demands of resource-constrained systems, ensuring minimal impact on device performance and battery life} \cite{tewari17ads}.

        While many efforts have been made in the literature to explore possible applications of CB algorithms and improve their accuracy, no previous study investigated approaches to reduce the computational complexity and the requirements of LinearUCB algorithms for implementation on edge devices with low-power and real-time requirements. This paper aims to fill this gap by exploring and developing software and hardware approaches that specifically reduce CB computational load and resource demands, making them more feasible and efficient for use in resource-constrained edge computing environments.

%% file: Sections/3.Algorithmic_Optimizations.tex
    \section{Algorithmic Optimizations}\label{Algorithmic_optimizations}
    
    The LinearUCB algorithms rely on the linearity assumption between the context and the reward, employing ridge regression procedures to assess the expected reward for each action and determine the optimal one.
    
    The key point of this process is matrix inversion, a computationally intensive operation whose complexity grows cubically with the size of the matrix.
    The Disjoint LinearUCB requires inverting $N$ matrices of dimensions $d \times d$ at each step, for an asymptotic complexity equal to $\mathcal{O}(d^3)$. The Hybrid variant is even more demanding, requiring the inversion of a large $k \times k$ and $N\ d \times d$ matrices, asymptotically growing as $\mathcal{O}(k^3)=\mathcal{O}(f^3d^3)$. This operation, common to both algorithmic variants, imposes a heavy computational load, resulting in performance bottlenecks and significant challenges for real-time applications where rapid decision-making is crucial. 
    To mitigate this problem, the approach in \cite{li2010contextual} proposes to update the original matrices at each step while computing and caching the inverses periodically instead of in real-time.
    However, this approach may limit online learning capabilities due to less frequent model updates, requires more memory and still necessitates longer iterations at regular intervals.

    Given that matrix inversion constitutes a bottleneck affecting the scalability and efficiency of LinearUCB algorithms, this section describes and explores mathematical strategies that can yield the same results without relying on this operation. First, we discuss the modifications applied to the simpler Disjoint LinearUCB algorithm, where the same principle has been adopted in previous works \cite{zr_obp, contextualbandits, ciucanu2022implementing, mann2016adaptive, guo2021survey}, and then we extend the approach to the more complex Hybrid model. For each variant, we thoroughly investigate the impact of the proposed techniques on the computational complexity, execution time, energy consumption, and memory requirements, which are essential aspects for applications on embedded systems, while varying the problem parameters.

    \subsection{Optimization of the Disjoint LinearUCB through the Sherman-Morrison Formula}  
    In the Disjoint algorithm, the $\mathbf{A_a}$ matrices are utilized in their inverted form, $\mathbf{A_a^{-1}}$, during action selection and subsequently adjusted in the normal form during the update phase based on the received reward. However, the process of updating and then inverting the matrix in each iteration introduces computational inefficiencies, especially as the matrix size increases.
        \begin{figure}[!t]
    \centering
    \Description{Optimized Disjoint LinearUCB algorithm. Matrix inversions are replaced using incremental updates based on the Sherman-Morrison formula.}
    \resizebox{0.53\textwidth}{!}{%
    \begin{minipage}{0.55\textwidth}
        \begin{algorithm}[H]
            \caption{Optimized Disjoint LinearUCB}\label{Algorithm:disj_SM}
            \begin{algorithmic}[1]
                \STATE \textbf{Input}\text{: $\alpha \in \mathbb{R}_{+}$ }
                \STATE \textbf{for} \text{t $= 1, 2, ...$, T}
                \STATE   \text{\ \ \ Observe the context for all arms $a\in \mathcal{A}_t$: $\mathbf{x}_{t,a} \in \mathbb{R}^{d}$ }
                \STATE \textbf{\ \ \ for all} \text{a $\in\mathcal{A}_t$}
                \STATE   \text{\ \ \ \ \ \ \textbf{if} t==1:}
                \STATE   \text{\ \ \ \ \ \ \ \ \ $\mathbf{A}_a^{\scriptscriptstyle -1} = \mathbf{I}_{d}$ }
                \STATE   \text{\ \ \ \ \ \ \ \ \ $\mathbf{b}_{a} = \mathbf{0}_{d \times 1}$}
                \STATE   \text{\ \ \ \ \ \ \textbf{endif}}
                \STATE   \text{\ \ \ \ \ \  $\hat{\boldsymbol{\theta}}_{a} = \mathbf{A}^{-1}_{a} \mathbf{b}_{a}$ }
                \STATE   \text{\ \ \ \ \ \  $p_{t, a} = \hat{\boldsymbol{\theta}}_{a}^{\top} \mathbf{x}_{t, a}+\alpha \sqrt{\mathbf{x}_{t, a}^{\top} \mathbf{A}^{-1}_{a} \mathbf{x}_{t,a}}$ }
                \STATE   \text{\ \ \ \textbf{endfor}}
                \STATE   \text{\ \ \ Choose $a_{t}=\arg \max _{a \in \mathcal{A}_{t}} (p_{t, a})$}
                \STATE   \text{\ \ \ and observe a real-valued payoff $r_t$}
                \STATE   \text{\ \ \ $\mathbf{A}^{-1}_{a_{t}} = 
                \mathbf{A}^{-1}_{a}-\frac{\mathbf{A}^{-1}_{a} \mathbf{x}_{t, a} \mathbf{x}_{t, a}^{\top} \mathbf{A}^{-1}_{a}}{1+\mathbf{x}_{t, a}^{\top} \mathbf{A}^{-1}_{a} \mathbf{x}_{t, a}}$ }
                \STATE   \text{\ \ \ $\mathbf{b}_{a_{t}} = \mathbf{b}_{a_{t}}+r_{t} \mathbf{x}_{t, a_{t}}$   }
                \STATE \textbf{endfor}
            \end{algorithmic}
        \end{algorithm}\vspace{-2em}
    \end{minipage}
    }
\end{figure}

    Considering that $\mathbf{A_a}$ undergoes incremental updates through a rank-one perturbation, $\mathbf{x}_t\mathbf{x}_t^{\top}$, it is well-suited for the application of the Sherman-Morrison (SM) theorem \cite{sherman1950adjustment}. 
    The theorem represents an efficient method for updating the inverse of a matrix, declaring that if $\mathbf{A} \in \mathbb{R}^{n\times n}$ is an invertible square matrix with inverse $\mathbf{A^{-1}}$, and it undergoes a rank-one update $\mathbf{uv}^{\top}$, with $\mathbf{u}, \mathbf{v} \in \mathbb{R}^{n}$, then the updated inverse can be efficiently computed by leveraging and updating the previous one, as demonstrated in (\ref{eq: Sherman-Morrison}).
    
    \begin{equation}\label{eq: Sherman-Morrison}
        \left(\mathbf{A}+\mathbf{u v}^{\top}\right)^{-1}=\mathbf{A}^{-1}-\frac{\mathbf{A}^{-1} \mathbf{u v}^{\top} \mathbf{A}^{-1}}{1+\mathbf{v}^{\top} \mathbf{A}^{-1} \mathbf{u}}
    \end{equation}

    The update rule in (\ref{eq:update_disj}) modified with the Sherman-Morrison formula for the Disjoint algorithm is shown in (\ref{eq:disj_upd_SM}), while Algorithm \ref{Algorithm:disj_SM} presents the complete optimized algorithm.
    \begin{equation}\label{eq:disj_upd_SM}
        \left(\mathbf{A}_{a_t}+\mathbf{x}_{t, a_t} \mathbf{x}_{t, a_t}^{\top}\right)^{-1}=\mathbf{A}_{a_t}^{-1}-\frac{\mathbf{A}_{a_t}^{-1} \mathbf{x}_{t, a_t} \mathbf{x}_{t, a_t}^{\top}  \mathbf{A}_{a_t}^{-1}}{1+\mathbf{x}_{t, a_t}^{\top} \mathbf{A}_{a_t}^{-1} \mathbf{x}_{t, a_t}}
    \end{equation}
    This approach enables the direct storage of the inverse matrix $\mathbf{A_a^{-1}}$, \conditionalhighlight{reducing the stack memory requirements for intermediate variables, and completely eliminates the matrix inversion operation,} significantly reducing the computational overhead by saving the inversion of $N [d \times d ]$ matrices at each time step, at the cost of a minimal increase in the complexity of the update phase.  
    
\begin{figure}[!b]
        \centering
        \Description{The proposed optimization reduces the asymptotic complexity of the Disjoint LinearUCB algorithm from $\mathcal{O}(d^3)$ to $\mathcal{O}(d^2)$.}
        \includegraphics[width=0.55\columnwidth]{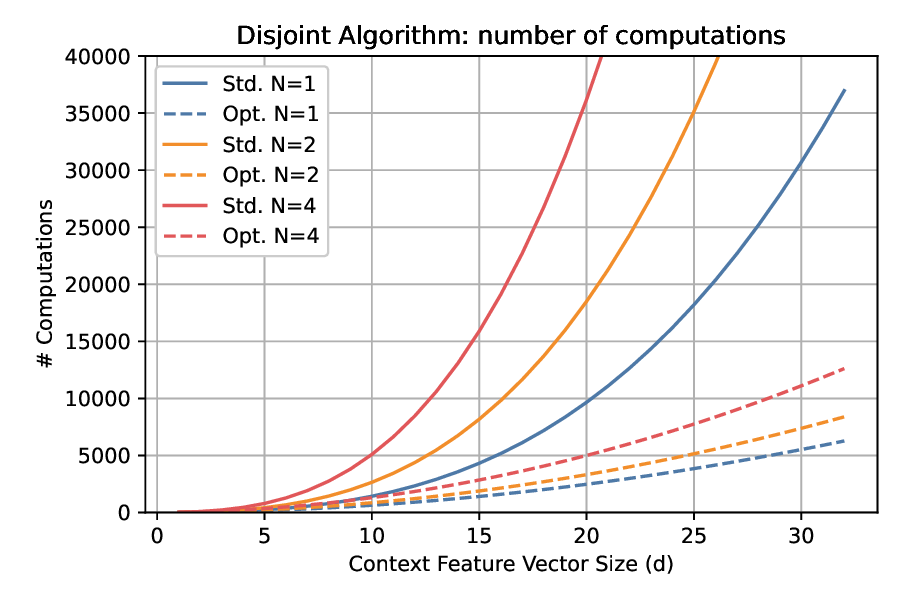}
        \caption{Computational complexity trends for the standard Disjoint Algorithm and the proposed optimized version.}
        \label{fig: Disjoint_complexity}
    \end{figure}

    Fig. \ref{fig: Disjoint_complexity} illustrates the computational complexity trends of the Disjoint algorithm varying \( N \) and \( d \) parameters, comparing the standard and optimized versions. The proposed optimization reduces the asymptotic complexity of the Disjoint algorithm from $\mathcal{O}(d^3)$ to $\mathcal{O}(d^2)$.

    \begin{figure}[!b]
        \centering
        \includegraphics[width=0.52\columnwidth]{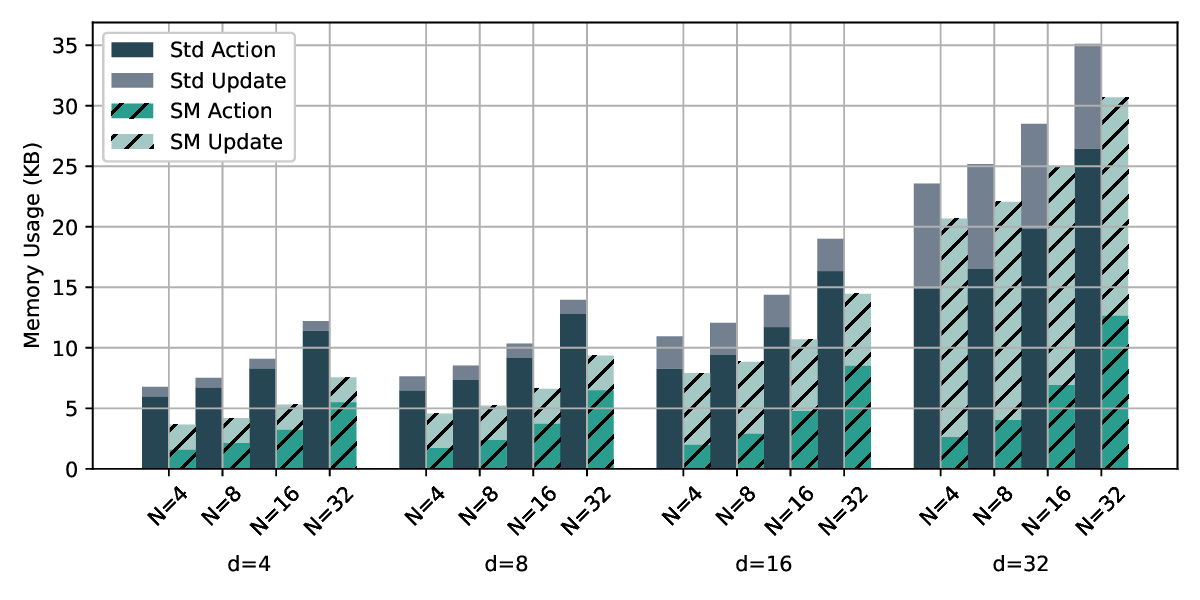}
        \caption{\conditionalhighlight{Stack Memory Usage for the standard Disjoint Algorithm and the proposed optimized version.}}
        \Description{The stack memory usage for the Disjoint LinearUCB algorithm during the action selection and update phases, obtained using a Python memory profiler. The plot shows the memory usage for both the standard and optimized versions while varying \( d \) and \( N \) from 4 to 32. The optimization reduces peak stack memory usage by up to 5 KB, shifting the memory demand from the action selection phase to the update phase. This improvement ensures feasibility for memory-constrained devices.}
        \label{fig: Disjoint_memory}
    \end{figure}
    
    Fig. \ref{fig: Disjoint_memory} \conditionalhighlight{depicts the stack memory usage, obtained through a Python memory profiler, of both the traditional and optimized versions during the action selection and update phases, with $d$ and $N$ varying from $4$ to $32$. This approach reduces peak stack memory usage by up to 5 KB, making it feasible to run the algorithms on memory-constrained devices. This analysis also reveals that the optimization shifts stack memory usage from the action selection phase to the update one.}

    \subsection{Optimization of the Hybrid LinearUCB through the Woodbury's Formula}
    In this section, the optimization strategies applied to the Disjoint algorithm are generalized and extended to the more complex Hybrid LinearUCB model.
    
    Similar to the Disjoint variant, each action in the Hybrid algorithm is associated with a $d \times d$ matrix $\mathbf{A}_a$, used in its inverted form during the action selection phase and incrementally updated via rank-one perturbations, as per (\ref{eq:hyb_upd2}). Consequently, the Sherman-Morrison formula is again applicable, allowing to directly store $\mathbf{A}_{a}^{-1}$.
    However, although this approach improves computational efficiency by removing $N [d\times d]$ matrix inversions at each iteration, the overall performance gain for this model remains modest. The computational complexity is, in fact, predominantly driven by the inversion of the large $k \times k$ shared matrix, $\mathbf{A}_0$.
    To illustrate, when considering $ d = 8 $ and $ f = 16 $, the inversion of $\mathbf{A}_0$ requires approximately $2,097,152$ operations per iteration, which is significantly higher than the relatively minor cost of $512$ operations for each $\mathbf{A}_0$ inversion. 
    Consequently, it is crucial to explore whether a similar optimization strategy can be applied to $\mathbf{A}_0$, aiming to substantially accelerate the algorithm.

        \begin{figure}[!t]
        \centering
            \Description{Optimized Hybrid LinearUCB algorithm. Matrix inversions are replaced using incremental updates based on the Sherman-Morrison-Woodbury formula.}
        \resizebox{0.7\textwidth}{!}{%
        \begin{minipage}{0.7\textwidth}
            \begin{algorithm}[H]
            \caption{optimized Hybrid LinearUCB }\label{Algorithm:hyb_SMW}
            \begin{algorithmic}[1]
            \STATE \textbf{Input}\text{: $\alpha \in \mathbb{R}_{+}$ }
            \STATE \tikzmk{A} \text{$\mathbf{A}_0^{\scriptscriptstyle -1} = \mathbf{I}_{k}$ }
            \STATE \text{${b}_{0} = {0}_{k}$ }
            \STATE \textbf{for} \text{t = 1, 2, ..., T}
            \STATE   \text{\ \ \ Observe the context for all arms $a\in \mathcal{A}_t$: $(\mathbf{z}_{t,a},\mathbf{x}_{t,a}) \in \mathbb{R}^{k+d}$ }
            \STATE \text{\ \ \ $ \hat{\boldsymbol{\beta}} = \mathbf{A}_0^{\scriptscriptstyle -1} \mathbf{b}_{0}$}
            \STATE \textbf{\ \ \ for all} \text{$a \in \mathcal{A}_t$}
            \STATE   \text{\ \ \ \ \ \ \textbf{if} t==1:}
            \STATE    \text{\ \ \ \ \ \ \ \ \ $\mathbf{A}_a^{\scriptscriptstyle -1} = \mathbf{I}_{d}$  }
            \STATE   \text{\ \ \ \ \ \ \ \ \ $\mathbf{B}_{a} = \mathbf{0}_{d \times k}$}
            \STATE   \text{\ \ \ \ \ \ \ \ \ $\mathbf{b}_{a} = \mathbf{0}_{d \times 1}$}
            \STATE   \text{\ \ \ \ \ \ \textbf{endif}}
            \STATE   \text{\ \ \ \ \ \  $\hat{\boldsymbol{\theta}}_{a} = \mathbf{A}_a^{\scriptscriptstyle -1}\left(\mathbf{b}_{a}-\mathbf{B}_{a} \hat{\boldsymbol{\beta}}\right)$}
            \STATE   \text{\ \ \ \ \ \  $\begin{aligned}
                                        s_{t, a} &= \mathbf{z}_{t, a}^{\scriptscriptstyle \top} \mathbf{A}_0^{\scriptscriptstyle -1} \mathbf{z}_{t, a}+
                                                             -2 \mathbf{z}_{t, a}^{\scriptscriptstyle \top} \mathbf{A}_0^{\scriptscriptstyle -1} \mathbf{B}_{a}^{\scriptscriptstyle \top} \mathbf{A}_a^{\scriptscriptstyle -1} \mathbf{x}_{t, a}+ \\
                                                            &+ \mathbf{x}_{t, a}^{\scriptscriptstyle \top} \mathbf{A}_a^{\scriptscriptstyle -1} \mathbf{x}_{t, a}+{x}_{t, a}^{\scriptscriptstyle \top} \mathbf{A}_a^{\scriptscriptstyle -1} \mathbf{B}_{a} \mathbf{A}_0^{\scriptscriptstyle -1} \mathbf{B}_{a}^{\scriptscriptstyle \top} \mathbf{A}_a^{\scriptscriptstyle -1} \mathbf{x}_{t, a}
                                        \end{aligned}$}
            \STATE   \text{\ \ \ \ \ \  $p_{t, a} = \mathbf{z}_{t, a}^{\scriptscriptstyle \top} \hat{\boldsymbol{\beta}}+{x}_{t, a}^{\scriptscriptstyle \top} \hat{\boldsymbol{\theta}}_{a}+\alpha \sqrt{s_{t, a}}$ }
            \STATE   \text{\ \ \ \textbf{endfor}}
            \STATE   \text{\ \ \ Choose $a_{t}=\arg \max _{a \in \mathcal{A}_{t}} (p_{t, a})$}
            \STATE   \text{\ \ \ and observe a real-valued payoff $r_t$}
            \STATE   \text{\ \ \ $\mathbf{A}_{0}^{\scriptscriptstyle -1} = \mathbf{A}_{0}^{\scriptscriptstyle -1}-\mathbf{A}_{0}^{\scriptscriptstyle -1} \mathbf{B}_{a_t}^{\scriptscriptstyle \top}(\mathbf\mathbf{A}_{a_t} + \mathbf{B}_{a_t} \mathbf{A}_{0}^{\scriptscriptstyle -1} \mathbf{B}_{a_t}^{\scriptscriptstyle \top})^{\scriptscriptstyle -1} \mathbf{B}_{a_t} \mathbf{A}_{0}^{\scriptscriptstyle -1}$  }
            \STATE   \text{\ \ \ $\mathbf{b}_{0} = \mathbf{b}_{0}+\mathbf{B}_{a_{t}}^{\scriptscriptstyle \top} \mathbf{A}_{a_{t}}^{\scriptscriptstyle -1} \mathbf{b}_{a_{t}}$}
            \STATE   \text{\ \ \ $\mathbf{A}_{a_{t}}^{\scriptscriptstyle -1} = \mathbf{A}_{a_{t}}^{\scriptscriptstyle -1}-\frac{\mathbf{A}_{a_{t}}^{\scriptscriptstyle -1} \mathbf{x}_{t, a_{t}} \mathbf{x}_{t, a_{t}}^{\scriptscriptstyle \top} \mathbf{A}_{a_{t}}^{\scriptscriptstyle -1}}{1+\mathbf{x}_{t, a_{t}}^{\scriptscriptstyle \top} \mathbf{A}_{a_{t}}^{\scriptscriptstyle -1}\mathbf{x}_{t, a_{t}}}$ }
            \STATE   \text{\ \ \ $\mathbf{B}_{a_{t}} = \mathbf{B}_{a_{t}}+\mathbf{x}_{t, a_{t}} \mathbf{z}_{t, a_{t}}^{\scriptscriptstyle \top}$ }
            \STATE   \text{\ \ \ $\mathbf{b}_{a_{t}} = {b}_{a_{t}}+r_{t} \mathbf{x}_{t, a_{t}}$}
            \STATE   \text{\ \ \ $\mathbf{A}_0^{\scriptscriptstyle -1} = \mathbf{A}_0^{\scriptscriptstyle -1}-\frac{\mathbf{A}_0^{\scriptscriptstyle -1} \mathbf{z}_{t, a_t} \mathbf{z}_{t, a_t}^{\scriptscriptstyle \top} \mathbf{A}_0^{\scriptscriptstyle -1}}{1+\mathbf{z}_{t, a_t}^{\scriptscriptstyle \top} \mathbf{A}_0^{\scriptscriptstyle -1} \mathbf{z}_{t, a_t}}$ }
            \STATE   \text{\ \ \ $\mathbf{A}_{0}^{\scriptscriptstyle -1} = \mathbf{A}_{0}^{\scriptscriptstyle -1}+\mathbf{A}_{0}^{\scriptscriptstyle -1} \mathbf{B}_{a_t}^{\scriptscriptstyle \top}(\mathbf{A}_{a_t} - \mathbf{B}_{a_t} \mathbf{A}_{0}^{\scriptscriptstyle -1} \mathbf{B}_{a_t}^{\scriptscriptstyle \top})^{\scriptscriptstyle -1} \mathbf{B}_{a_t} \mathbf{A}_{0}^{\scriptscriptstyle -1}$  }
            \STATE   \text{\ \ \ $\mathbf{b}_{0} = \mathbf{b}_{0}+r_{t} \mathbf{z}_{t, a_{t}}-\mathbf{B}_{a_{t}}^{\scriptscriptstyle \top} \mathbf{A}_{a_{t}}^{\scriptscriptstyle -1} \mathbf{b}_{a_{t}}$}
            \STATE \textbf{endfor}
            \end{algorithmic}
        \end{algorithm}
    \end{minipage}
    }
\end{figure}
 
    The $\mathbf{A}_0$ matrix is always used in its inverted form during the action choice, and it is then incrementally updated according to (\ref{eq:hyb_upd1}) and (\ref{eq:hyb_upd2}), to which in the following we will refer as first and second update step, respectively.  

    In the first one, the Hybrid algorithm modifies $\mathbf{A}_0$ by incorporating the term $\mathbf{B}_{a_t}^{\top}\mathbf{A}_{a_t}^{-1}\mathbf{B}_{a_t}$, which represents a complex adjustment that goes beyond the normal rank-one perturbation, entailing the multiplication of three matrices.
    To efficiently address this more complex update scenario, we apply a generalized version of the Sherman-Morrison formula, known as the Woodbury matrix identity \cite{hager1989updating}. This formula is useful when the matrix update process involves an incremental rank-$k$ perturbation. 
    Given an invertible matrix $\mathbf{A} \in \mathbb{R}^{n\times n}$ with its inverse $\mathbf{A^{-1}}$, and matrices $\mathbf{U} \in \mathbb{R}^{n\times k}$, $\mathbf{C} \in \mathbb{R}^{k\times k}$, and $\mathbf{V} \in \mathbb{R}^{k\times n}$, the inverse of the updated matrix after applying a perturbation $\mathbf{UCV}$ can be computed by (\ref{eq:Woodbury}).
    \begin{equation}\label{eq:Woodbury}
    (\mathbf{A}+\mathbf{U C V})^{-1} = \mathbf{A}^{-1} - \mathbf{A}^{-1} \mathbf{U}(\mathbf{C}^{-1} + \mathbf{V A}^{-1} \mathbf{U})^{-1} \mathbf{V A}^{-1}
    \end{equation}
    Applying the equation to the Hybrid algorithm makes it possible to modify the first update step as shown in  (\ref{eq:hyb_upd1_W_0}). 
        \begin{equation}\label{eq:hyb_upd1_W_0}
        \begin{split}
            (\mathbf{A}_0 + \mathbf{B}_{a_t}^{\top} \mathbf{A}_{a_t}^{-1} \mathbf{B}_{a_t})^{-1} = & \ \mathbf{A}_0^{-1} - \mathbf{A}_0^{-1} \mathbf{B}_{a_t}^{\top}\cdot \\
            & \cdot (\mathbf{A}_{a_t}+\mathbf{B}_{a_t} \mathbf{A}_0^{-1} \mathbf{B}_{a_t}^{\top})^{-1} \mathbf{B}_{a_t} \mathbf{A}_0^{-1}   
        \end{split}
    \end{equation}
    \conditionalhighlight{Notably, $\mathbf{A}_{a_t}$ required in this equation can be derived by inverting $\mathbf{A}^{-1}_{a_t}$, consistent with the traditional algorithm, where  $\mathbf{A}^{-1}_{a_t}$ is calculated through inversion during the update phase.}

    The second update phase involves the sum of the outer product $\mathbf{z_{t,a_t}}\mathbf{z_{t,a_t}}^{\top}$ to $\mathbf{A}_0$ and the subtraction of $\mathbf{B}_{a_t}^{\top}\mathbf{A}_{a_t}^{-1}\mathbf{B}_{a_t}$. 

    The first sum involves a rank-one incremental update, since $\mathbf{z_{t,a_t}}$ is a column vector in $\mathbb{R}^{k}$. Thus, this step can be efficiently handled using the Sherman-Morrison formula, as depicted in (\ref{eq:hyb_upd2.1_SM}).
    \begin{equation}\label{eq:hyb_upd2.1_SM}
        (\mathbf{A}_{0}+\mathbf{z}_{t, a_t} \mathbf{z}_{t, a_t}^{\top})^{-1}=\mathbf{A}_{0}^{-1}-\frac{\mathbf{A}_{0}^{-1} \mathbf{z}_{t, a_t} \mathbf{z}_{t, a_t}^{\top} \mathbf{A}_{0}^{-1}}{1+\mathbf{z}_{t, a_t}^{\top} \mathbf{A}_{0}^{-1} \mathbf{z}_{t, a_t}}
    \end{equation}

    The subtraction, instead, mirrors the mechanics of the first update step. The modified update rule is shown in (\ref{eq:hyb_upd2_W}).
    \begin{equation}\label{eq:hyb_upd2_W}
        \begin{split}
            (\mathbf{A}_0 - \mathbf{B}_{a_t}^{\top} \mathbf{A}_{a_t}^{-1} B_a)^{-1} = & \ \mathbf{A}_0^{-1} + \mathbf{A}_0^{-1} \mathbf{B}_{a_t}^{\top}\cdot \\
            & \cdot (\mathbf{A}_{a_t}-\mathbf{B}_{a_t} \mathbf{A}_0^{-1} \mathbf{B}_{a_t}^{\top})^{-1} \mathbf{B}_{a_t} \mathbf{A}_0^{-1}   
        \end{split}
    \end{equation}

    \begin{figure}[!t]
        \centering
        \Description{Computational complexity trends for the standard Hybrid Algorithm and the proposed optimized version. The optimization leads to an asymptotic computational complexity of $\mathcal{O}(k^2d)$ or $\mathcal{O}(f^2d^3)$ instead of the larger $\mathcal{O}(k^3)$ or $\mathcal{O}(f^3d^3)$}
        \includegraphics[width=0.6\columnwidth]{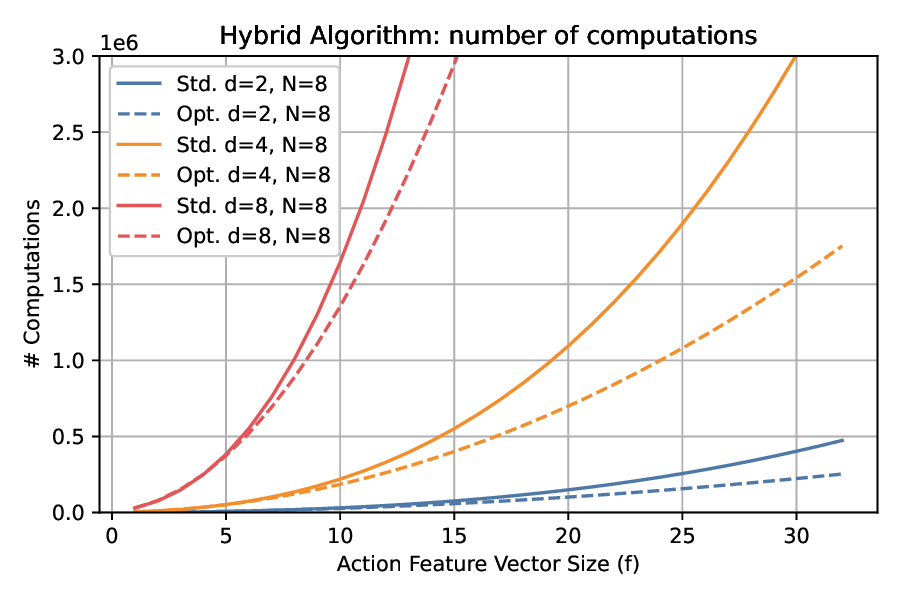}
        \caption{Computational complexity trends for the standard Hybrid Algorithm and the proposed optimized version.}
        \label{fig: Hybrid_complexity}
    \end{figure}

    Algorithm \ref{Algorithm:hyb_SMW} presents the optimized Hybrid algorithm.
    As depicted, this approach only involves the $\mathbf{A}_0^{-1}$ and $\mathbf{A}_a^{-1}$ matrices, updating them using the Sherman-Morrison-Woodbury (SMW) formulas. This enables the direct storage of the inverse matrices and \conditionalhighlight{reduces
    the number of matrix inversions from $N$ inversions of $[d \times d]$ matrices and one inversion of a $k \times k$ matrix per iteration to a constant $3[d \times d]$ inversions per iteration in the update process. This notably decreases computational complexity by removing the need for the large $k \times k$ inversion and eliminating the dependency on $N$.} 

    Fig. \ref{fig: Hybrid_complexity} depicts the computational complexity trends of the Hybrid algorithm with a constant \( N = 8 \), while varying \( f \), and \( d \), contrasting the standard and optimized versions. Overall, the optimization leads to an asymptotic computational complexity of $\mathcal{O}(k^2d)$ or $\mathcal{O}(f^2d^3)$ instead of the larger $\mathcal{O}(k^3)$ or $\mathcal{O}(f^3d^3)$. These results also illustrate the relative complexity of the Hybrid algorithm compared to the Disjoint variant, showing how rapidly the computational demands escalate with increasing problem parameters.

    Table \ref{table: complexity} summarizes the impact of the proposed algorithmic optimizations on the computational complexity of the algorithms, both during the action selection and update processes (Stage 1 and Stage 2).
    \begin{table}[!t]
        \centering
        \Description{Table resuming the asymptotic computational complexity of the LinearUCB algorithms. The proposed optimization reduces the complexity from $\mathcal{O}(d^3)$ and $\mathcal{O}(d^2)$ to $\mathcal{O}(k^3)$& $\mathcal{O}(k^2d)$ in the stage 1 and 2 of the Disjoint algorithm, respectively. The complexity is reduced from $\mathcal{O}(d^2)$ and $\mathcal{O}(d^2)$ to $\mathcal{O}(k^2d)$ and $\mathcal{O}(k^2d)$ in the stage 1 and 2 of the Hybrid algorithm, respectively.}
        \caption{Asymptotic computational complexity in the original Disjoint and Hybrid Algorithms and in the proposed optimized versions}
        \label{table: complexity}
        \begin{tabular}{l| l | c| c | c}  
             & \multicolumn{2}{c|}{\textbf{Disjoint LinearUCB}} & \multicolumn{2}{c}{\textbf{Hybrid LinearUCB}} \\
             & Stage 1 & Stage 2 & Stage 1 & Stage 2 \\\hline
            Standard & $\mathcal{O}(d^3)$& $\mathcal{O}(d^2)$& $\mathcal{O}(k^3)$& $\mathcal{O}(k^2d)$\\
            Optimized & $\mathcal{O}(d^2)$& $\mathcal{O}(d^2)$& $\mathcal{O}(k^2d)$& $\mathcal{O}(k^2d)$\\\hline        
        \end{tabular}
    \end{table}

\begin{figure}[!t]
        \centering
        \includegraphics[width=0.6\linewidth]{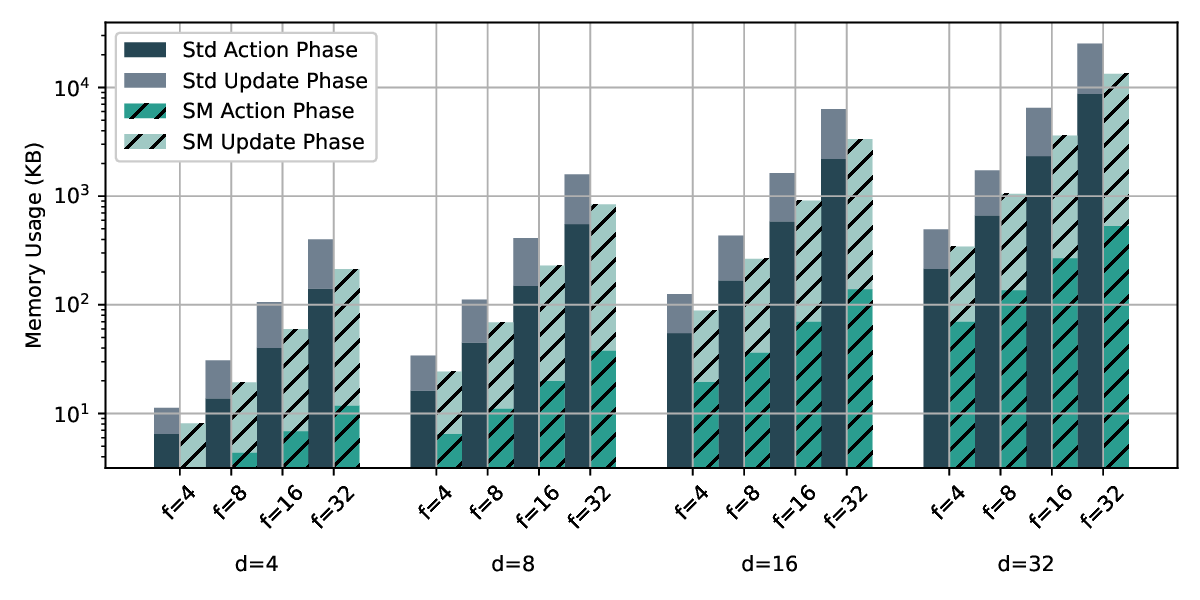}
        \caption{\conditionalhighlight{Stack Memory Usage for the standard Hybrid Algorithm and the proposed optimized version.}}
        \Description{The stack memory usage for the Hybrid LinearUCB algorithm, measured using a Python memory profiler. The plot compares the standard and optimized versions while varying \( d \) and \( f \) from 4 to 32. The optimization reduces peak memory usage by up to 12 KB, making the algorithm more suitable for deployment on embedded systems with limited memory resources.}
        \label{fig: Hybrid_memory}
    \end{figure}

    Fig. \ref{fig: Hybrid_memory} \conditionalhighlight{shows the stack memory usage of the hybrid algorithm, measured using a Python memory profiler, with the $d$ and $f$ parameters varying from $4$ to $32$. The proposed optimization reduces peak memory usage by up to 12 KB.}

    \subsection{Stability and Round-off Errors}
    In the previous section, we demonstrated how the Sherman–Morrison–Woodbury formula can significantly reduce the computational cost of the LinearUCB algorithm by avoiding complex and time-consuming matrix inversions. However, while this formula offers an exact solution for updating matrix inverses, its practical application can be affected by round-off errors due to finite-precision arithmetic, potentially leading to instabilities when the matrix becomes ill-conditioned.    
    In this section, we first establish the stability of the proposed approach by demonstrating that the inverse covariance matrices in the LinearUCB algorithms remain well-conditioned and stable over time. We then investigate the impact of round-off errors as the number of iterations grows, using synthetic datasets to quantify error accumulation. Finally, we present a novel strategy to mitigate these errors in scenarios where high precision is critical. We begin our analysis with the Disjoint LinearUCB algorithm, addressing its simpler structure, and then we extend our findings to the more complex Hybrid one.
    
    \subsubsection{Stability of the Covariance Matrices}

    Round-off or rounding errors arise from the finite-precision arithmetic inherent in digital computations and can accumulate over successive operations, potentially leading to numerical inaccuracies. When recomputing the inverse of a matrix from scratch after each modification, each inversion is independent of previous computations, making this approach generally more resilient to the accumulation of round-off errors. In contrast, in incremental update mechanisms, small errors can accumulate across iterations, which may gradually lead to noticeable inaccuracies. In detail, in the SM formula, round-off errors are prone to amplification and can introduce instabilities in two main scenarios:
    \begin{enumerate}
    \item Denominator Approaching Zero: The SM formula contains a denominator term that, when approaching zero, can render the update unstable. This scenario can lead to large numerical errors or even division by zero, resulting in computational failures.
    
    \item Singular or Ill-Conditioned Inverse Matrices: If the inverse matrix becomes singular or ill-conditioned the calculations may amplify numerical errors. An ill-conditioned matrix is highly sensitive to minor perturbations, which can result in substantial deviations in the output, thereby compromising the algorithm’s stability.
    \end{enumerate}
    To ensure the stability of the SM update in our optimized LinearUCB algorithms, in this section we demonstrate that these two conditions are avoided. To do that, is sufficient to show that the involved covariance matrices  $\mathbf{A}_{a_{t}}$ remain positive and definite throughout the iterations \cite{Kalman}. This is due to two key reasons:
    
    \begin{enumerate}

        \item If \( \mathbf{A}_{a_t} \) is positive definite, then the updated matrix \( \mathbf{A}_{a_{t+1}} = \mathbf{A}_{a_{t}} + \mathbf{x}_t \mathbf{x}_t^\top \) and its inverse \( \mathbf{A}^{-1}_{a_{t+1}} \) are also positive definite ensuring that the condition number of \( \mathbf{A}_{a_{t}} \) remains bounded, preventing the matrix from becoming ill-conditioned.
        In fact, if \( \mathbf{A}_{a_{t}} \) is positive definite at iteration \( t \), since \( \mathbf{x}_t \mathbf{x}_t^\top \) is positive semi-definite for any non-zero \( \mathbf{x}_t \), the sum of a positive definite matrix \( \mathbf{A}_{a_{t}} \) and a positive semi-definite matrix \( \mathbf{x}_t \mathbf{x}_t^\top \) remains positive definite.
    
        \item if the matrix is positive definite, the denominator in the Sherman-Morrison formula, \( d_t = 1 + \mathbf{x}_t^\top \mathbf{A}_{a_{t}}^{-1} \mathbf{x}_t \), is always strictly greater than one.
        In fact, since \( \mathbf{A}_{a_{t}}^{-1} \) is positive definite (as proven in the previous step), for any non-zero \( \mathbf{x}_t \), we have \(\mathbf{x}_t^\top \mathbf{A}_{a_{t}}^{-1} \mathbf{x}_t > 0\). Therefore, the denominator \( d_t = 1 + \mathbf{x}_t^\top \mathbf{A}_{a_{t}}^{-1} \mathbf{x}_t > 1 \).
    \end{enumerate}

    In the optimized Disjoint LinearUCB algorithm proposed in Algorithm \ref{Algorithm:disj_SM}, the inverse covariance matrices \( \mathbf{A}_a^{-1} \) are positive definite by construction, as they are initialized as \( \mathbf{A}_{a} = \lambda \mathbf{I}_d \) at time-step \( t = 0 \), where \( \lambda > 0 \) is the regularization parameter and \( \mathbf{I}_d \) is the \( d \times d \) identity matrix. From what we demonstrated, this implies that the denominator in the SM formula remains greater than one and that these matrices remain positive definite and well-conditioned throughout the iterations. As a result, round-off errors are not amplified, and the algorithm is expected to maintain numerical stability over time. 

   The same analysis extends to the more complex optimized Hybrid LinearUCB algorithm presented in Algorithm \ref{Algorithm:hyb_SMW}, which incorporates global features shared across all arms. We already proved that each \( \mathbf{A}_{a_{t}} \) matrix remains positive definite after the update phase; for the Hybrid case, we now demonstrate the same for the global covariance matrix \( \mathbf{A}_0 \) and its update. At time-step \( t = 0 \), \( \mathbf{A}_0 \) is initialized as \( \lambda \mathbf{I}_k \), where \( \lambda > 0 \) and \( \mathbf{I}_k \) is the \( k \times k \) identity matrix. At each iteration \( t \), \( \mathbf{A}_0 \) is updated when an arm \( a_t \) is selected according to (\ref{eq:hyb_upd1}) and (\ref{eq:hyb_upd2}). In (\ref{eq:hyb_upd1}), \( \mathbf{z}_{t,a_t} \mathbf{z}_{t,a_t}^\top \) is positive semi-definite (as it is an outer product of a vector with itself), and thus, adding it to the positive definite \( \mathbf{A}_0 \) results in a positive definite matrix. The resulting matrix is further updated with the term \( \mathbf{B}_{a_t}^\top \mathbf{A}_{a_t}^{-1} \mathbf{B}_{a_t} \) in (\ref{eq:hyb_upd2}). To show that this term is positive semi-definite, we need to demonstrate that for any vector \( \mathbf{x} \),
    \[
    \mathbf{x}^\top (\mathbf{B}_{a_t}^\top \mathbf{A}_{a_t}^{-1} \mathbf{B}_{a_t}) \mathbf{x} \geq 0.
    \]
    
    However, since we already demonstrated that the matrix \( \mathbf{A}_{a_t}^{-1} \) is positive definite, we know that for any non-zero vector \( \mathbf{y} \), \( \mathbf{y}^\top \mathbf{A}_{a_t}^{-1} \mathbf{y} > 0 \). Setting \( \mathbf{y} = \mathbf{B}_{a_t} \mathbf{x} \) gives:
    \[
    \mathbf{x}^\top \mathbf{B}_{a_t} \mathbf{A}_{a_t}^{-1} \mathbf{B}_{a_t}^\top \mathbf{x} = \mathbf{y}^\top \mathbf{A}_{a_t}^{-1} \mathbf{y} \geq 0.
    \]
    This confirms that \( \mathbf{B}_{a_t} \mathbf{A}_{a_t}^{-1} \mathbf{B}_{a_t}^\top \) is positive semi-definite, and therefore, \( \mathbf{A}_0 \) remains positive definite.

    Consequently, both the global and arm-specific covariance matrices in the Hybrid algorithm remain positive definite over time. 
    
    By proving that the matrices of the algorithms are positive definite, we ensure that their inverses are also positive definite and well-conditioned. This positive definiteness of the inverses is crucial for maintaining numerical stability in the SMW updates, as it prevents the amplification of round-off errors and ensures the robustness of the algorithms over time. Thus, the optimized LinearUCB algorithm is robust against numerical issues associated with finite-precision arithmetic.
    
    \subsubsection{Assessing the Accumulation of Round-Off Errors}
    In the previous section, we demonstrated that the covariance matrices within the LinearUCB algorithms remain well-conditioned throughout the iterations, making them suitable candidates for the Sherman–Morrison update. This condition prevents the amplification of round-off errors over time, allowing the incremental update approach in the optimized LinearUCB algorithms to maintain numerical stability effectively.

\begin{figure}[!ht]
        \centering
        \includegraphics[width=\textwidth]{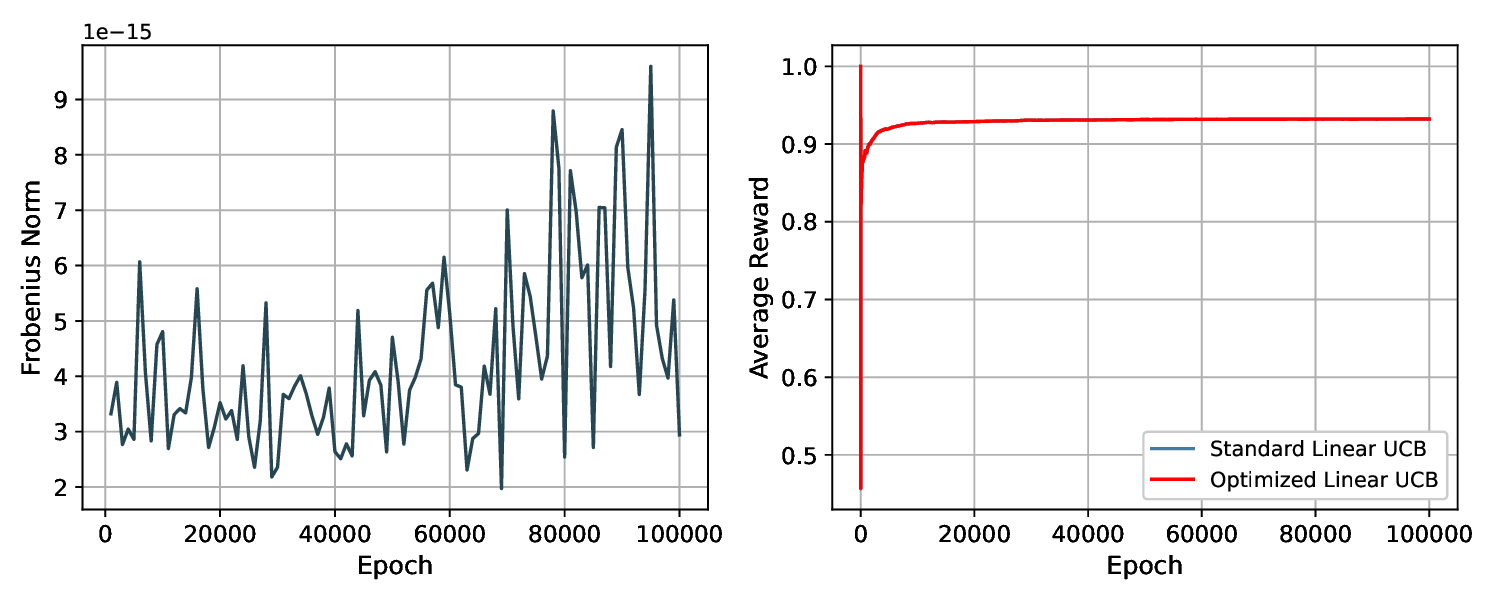}
        \caption{Error growth in the Disjoint Linear UCB algorithm, measured by the Frobenius norm of the difference between \( \mathbf{A}_{a_t}^{-1} \) (computed incrementally) and the same matrix obtained via full inversion.}
        \Description{In the Disjoint Linear UCB algorithm, the error growth, measured using the Frobenius norm, remains consistently around \(1 \times 10^{-15}\) over 100,000 iterations. This indicates that round-off errors accumulate very slowly and have a negligible impact on the algorithm's numerical stability.}
        \label{fig:disjoint_errors}
    \end{figure}

\begin{figure}[!ht]
        \centering
        \includegraphics[width=\textwidth]{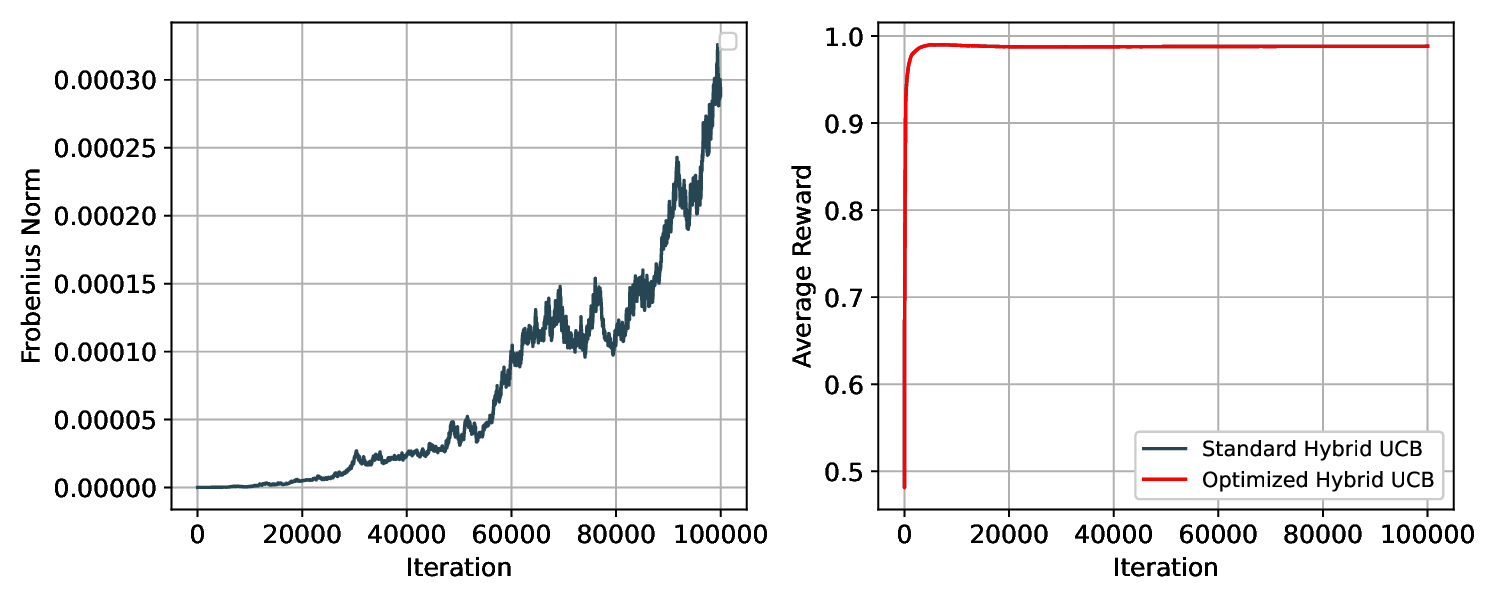}
        \caption{Error growth in the Hybrid Linear UCB algorithm, measured by the Frobenius norm of the difference between \( \mathbf{A}_{0}^{-1} \) (computed incrementally) and the same matrix obtained via full inversion.}
        \Description{In the Hybrid Linear UCB algorithm, the error growth in the global matrix \( \mathbf{A}_0 \) increases to \(1 \times 10^{-4}\) over 100,000 iterations. This larger accumulation is attributed to the frequent updates of \( \mathbf{A}_0 \) at each iteration, making it more prone to round-off error accumulation compared to arm-specific matrices.}
        \label{fig:hybrid_errors}
    \end{figure}

    To further evaluate the practical impact of round-off errors, we conducted an empirical assessment. We ran 100,000 iterations on synthetic datasets for both the Disjoint and Hybrid LinearUCB algorithms. At each iteration, we compared the matrices \( \mathbf{A}_{a_{t}}^{-1} \) and \( \mathbf{A}_0^{-1} \) obtained through incremental updates with those obtained via the traditional full matrix inversion approach. Figures \ref{fig:disjoint_errors} and \ref{fig:hybrid_errors} display the error growth, computed as the Frobenius norm of the difference between matrices, and the average reward for both the Disjoint and Hybrid LinearUCB algorithms.
    
    For the Disjoint algorithm, the total difference measured using the Frobenius norm, consistently hovers around \( 1 \times 10^{-15} \), indicating that round-off errors accumulate very slowly and remain completely negligible even after 100000 iterations. 
    
    In contrast, in the Hybrid Linear UCB algorithm, we observed a larger accumulation of round-off errors, particularly in the global matrix \( \mathbf{A}_0 \), where the error grew up to \( 1 \times 10^{-4} \) after 100000 iterations. This larger accumulation is primarily due to the fact that while the arm-specific matrices \( \mathbf{A}_a \) are updated only when a particular arm is selected, the global matrix \( \mathbf{A}_0 \) is updated at every iteration, making it more prone to round-off error accumulation.
    
    An important observation is that in both algorithms, the optimization introduced by the Sherman-Morrison-Woodbury approach had no effect on the decision-making process. Across 100,000 iterations, the actions selected by the optimized methods were identical to those chosen by the traditional matrix inversion, resulting in the same cumulative reward for both approaches. This suggests that the round-off errors introduced by the incremental updates had a negligible impact on the algorithm’s performance, even after a substantial number of iterations.

    \subsubsection{Proposed Solution to Mitigate Round-Off Errors}
        \begin{figure}[!t]
        \centering
        \includegraphics[width=\textwidth]{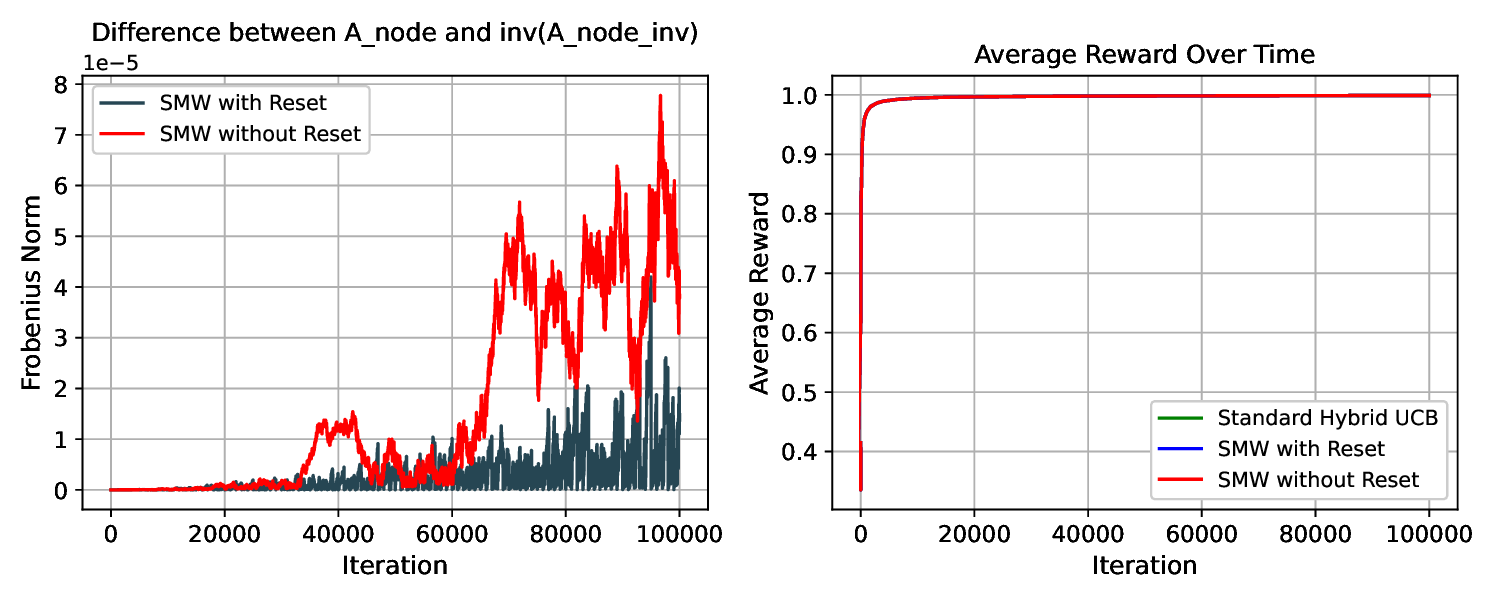}
        \caption{Error growth in the Hybrid Linear UCB algorithm with periodic correction of the inverse matrices (every 5000 iterations).}
        \Description{By applying a periodic correction strategy in the Hybrid Linear UCB algorithm, where the inverse matrices are recalculated every 5,000 iterations, the total round-off error is reduced by up to two orders of magnitude. This demonstrates the effectiveness of periodic corrections in mitigating error accumulation while preserving computational efficiency.}
        \label{fig:hybrid_corrected_errors}
    \end{figure}

    While our results show that the round-off errors in the optimized SMW approach have negligible impact on performance, in certain applications where the algorithm may be run for an extremely high number of iterations, it could be important to mitigate the accumulation of these errors. To address this, we propose a method to periodically correct the inverse matrices, thus alleviating the influence of round-off errors over time.    
    This approach consists of periodically replacing or correcting the inverse matrices \( \mathbf{A}_a^{-1} \) and \( \mathbf{A}_0^{-1} \). When memory constraints are not a major issue, we can store both the direct matrices \( \mathbf{A}_a \) and \( \mathbf{A}_0 \), as well as their inverses. The direct matrices are updated in each iteration using the traditional algorithmic formulation (i.e., without incremental updates), while the inverse matrices are incrementally updated via the SMW approach and used for action selection.
    Every \( N \) iterations, where \( N \) is a tunable integer, the inverse matrices \( \mathbf{A}_a^{-1} \) and \( \mathbf{A}_0^{-1} \) are replaced by the exact inverses computed from their corresponding direct matrices. This periodic correction reduces the impact of round-off errors without requiring matrix inversion at every step, thus maintaining the computational efficiency of the SMW approach.
    The trade-off here is between the complexity of periodically computing matrix inversions and the degree to which round-off errors are mitigated. The value of \( N \) can be adjusted depending on the specific application requirements, balancing between minimizing error growth and minimizing the overhead introduced by performing full matrix inversions.
    
    To demonstrate the effectiveness of this approach, we implemented this mechanism in the Hybrid Linear UCB algorithm and ran it for 100,000 iterations with \( N = 5000 \). The results, shown in Figure \ref{fig:hybrid_corrected_errors}, indicate that the total round-off error can be reduced by up to two orders of magnitude compared to the original uncorrected approach.

    This periodic correction approach provides a solution for applications where the algorithm is expected to run for an extremely high number of iterations, and the accumulation of round-off errors could eventually affect performance. By only computing full matrix inversions every \( N \) iterations, this method strikes a balance between computational efficiency and numerical precision. However, it introduces additional memory requirements and complexity. Notably, the overall computational load remains significantly reduced compared to traditional approaches, as matrix inversion is performed only once every \( N \) iterations, as opposed to every iteration. The value of \( N \) thus provides a flexible parameter to trade off between complexity and round-off error mitigation.

%% file: Sections/4.Hardware_Optimizations.tex
    \section{Vector Computing Acceleration}\label{Hardware Acceleration}
    Vector hardware accelerators are specialized computing units designed to efficiently process operations on vectors of numbers, exploiting data-level parallelism \cite{general_vector_processor}. While vector processors have been traditionally employed in supercomputers, there is an evident trend in adopting vector computing support in embedded processing systems \cite{barbirotta2023fault, 9232940}. 
    Given the strong dependence of LinearUCB algorithms on matrix operations, discussed in Sections \ref{Background_&_Relatedworks} and \ref{Algorithmic_optimizations}, this Section explores if and how vector computing architectural support may be harnessed to improve the algorithm execution time by implementing fundamental matrix operations in terms of elementary operations on vectors.
        
    \subsection{Reference Embedded Vector Computing Platform}
    For the purpose of our study, without loss of generality we adopted the Klessydra embedded processor family \cite{Klessydra-T13, dynamicdmr, barbirotta2023fault} to explore the impact of vector computing acceleration on the execution of the proposed algorithm. Klessydra processors are compliant with the open RISC-V instruction set architecture and are fully compatible with the PULPino open-source System-on-Chip platform. 
    
    The Klessydra-T13 core features a Vector Computing Unit (VCU) developed as a configurable hardware accelerator specifically designed for vector operations. The unit is equipped with dedicated local data storage referred to as Scratchpad Memories (SPMs) \conditionalhighlight{and can be configured at synthesis-time in terms of the SIMD (Single Instruction Multiple Data), which defines the number of vector elements processed in parallel}. A custom RISC-V-compliant instruction set extension, integrated in the RISC-V GCC compiler tool-chain, facilitates the programmer's access to the VCU by exposing elementary vector arithmetic operations as intrinsic function calls. \conditionalhighlight{The vector size is dynamically specified at runtime using a dedicated Current Status Register. The VCU leverages a hardware loop mechanism to efficiently repeat the required instruction until all vector elements are processed—based on the configured SIMD and vector size—without requiring additional instruction fetches.} The vector operations used for matrix-related computations in the algorithms of the proposed study are reported in the following section. 
    
    To implement and execute LinearUCB algorithms on the Klessydra-T13 embedded processor, constrained by lack of dynamic memory allocation and of floating-point support, we developed a versatile custom C++ library based on static allocation and fixed-point data representation. In order to quantify the impact of vector computing support on the final performance, the library was designed to process matrix operations in two distinct modes: a software-only mode, compiled on the standard RISC-V instruction set executed by the scalar processing core, and one vectorized version utilizing the VCU support. In the vectorized version, each arithmetic matrix operation was meticulously parallelized using the specialized vector intrinsic functions supported by the compiler toolchain.

    \subsection{Mapping LinearUCB algorithms on the vector accelerator}
    
    The application of the VCU specialized functions to Algorithms \ref{Algorithm:disj_SM} and \ref{Algorithm:hyb_SMW} is detailed below.
    \begin{itemize}
        \item matrix addition: this algorithm step was encoded using the \emph{kaddv (rd), (rs1), (rs2)} instruction, which allows the addition of two vectors located in the SPMs in a SIMD fashion. Executing this operation for each row, reduces the matrix addition time by a factor equal to the number of parallel functional units in the VCU.
        \item matrix multiplication: this algorithm step was encoded using the \emph{kvmulpsrf (rd), (rs1), (rs2)} instruction that performs a fixed-point multiplication between a vector from the scratchpad memory and a scalar from the register file. We used an iterative approach in which an entire row of the first matrix is multiplied by a scalar element from a second row.    This produces intermediate products, which are accumulated across vector lanes to form the corresponding row elements of the resultant matrix. By repeating this process for each row, the ability to perform multiple vector-scalar multiplications in a single cycle is fully leveraged, significantly enhancing the operation speed. The vector-scalar product is executed in hardware in double integer precision, and the resulting product is subsequently right-shifted by the appropriate scaling factor to maintain the output data width.     
        \item matrix inversion: this algorithm step was implemented by applying the LU factorization technique, which was encoded exploiting the \textit{kvdiv (rd), (rs1), (rs2)} instruction for executing element-wise vector divisions \cite{VEC_VLNPD} leveraging the variable latency dividers integrated in the VCU \cite{VLNPD}. 
    \end{itemize}
    
\begin{figure}[!t]
        \centering
        \Description{The vector hardware acceleration achieves a maximum speedup of 4, 29 and 15 times for the sum, multiplication and inversion operations, respectively}
        \includegraphics[width=0.55\columnwidth]{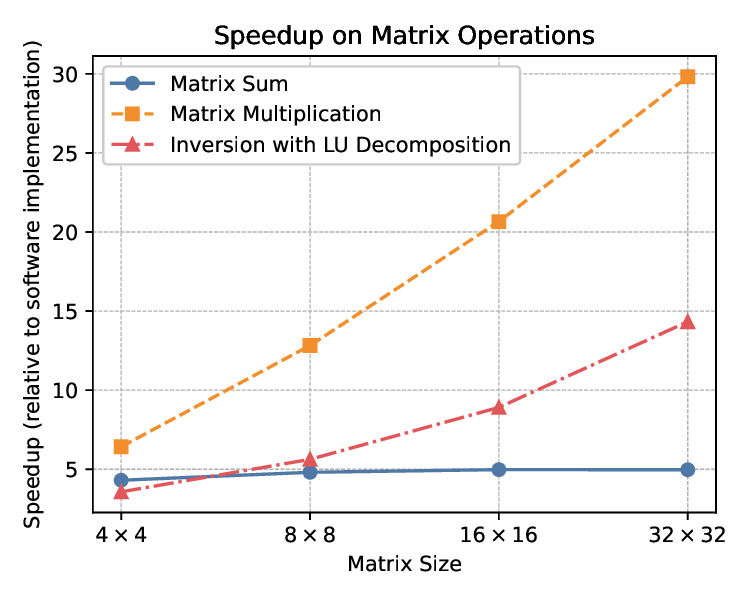}
        \caption{Speedup attained by the Klessydra-T13 vector computing acceleration on the fundamental matrix computation kernels used in LinearUCB algorithms}
        \label{fig: Matrix_operations}
    \end{figure}

    Figure \ref{fig: Matrix_operations} presents the performance gains achieved for the above discussed matrix operations, using a SIMD equal to 4. The speedup metric was chosen as the ratio between the clock cycles required by the standard operation (i.e. compiled for the RISC-V instruction set and executed on the Klessydra-T13 processor without vector computing support) and the cycles required by the accelerated operation (i.e. compiled on the extended instruction set exploiting the vector computing support of the Klessydra-T13 core). The speedup result is analyzed across different matrix sizes.    
    The matrix addition exhibits maximum benefit from parallel processing, demonstrating a consistent speedup across all matrix sizes, equivalent to a factor of $4\times$. 
    Matrix multiplication displays a different speedup pattern as the matrix size grows. The adopted parallel approach reaches a speed up equal to $6.42\times$ in the case of $4\times 4$ matrices, which scales up to $29.82\times$ for $32\times 32$. The speedup factor greater than the SIMD amount is explained by the heavy memory access overhead suffered in the standard scalar execution.
    The inversion operation represents the most challenging operation to be parallelized owing to its inherently sequential nature. Nevertheless, utilizing LU matrix factorization techniques and the optimized dividers in the VCU allows a notable acceleration of up to $15\times$ for $32\times 32$ matrices. 
      
%% file: Sections/5.Results.tex
    \section{Results on full algorithm execution}\label{Results}
    This section analyzes the cumulative impact of the proposed dual optimization strategy on the LinearUCB algorithms implemented and executed on embedded processor platforms. \conditionalhighlight{The purpose of the reported performance data is to contextualize the results obtainable with algorithmic enhancement and explicit vectorization in the low-end embedded system domain. For this reason, we chose the Cortex M4 STM32 Nucleo board, the Cortex A72 Raspberry PI 4 Model B board, and the Klessydra T13 PULPino open-source soft-processor equipped with basic vector acceleration. The contextualization of the results does not aim to demonstrate absolute superior performance to specific commercial products, yet to give evidence of the effectiveness of the explored techniques in limited hardware resource scenarios.}
    
    For the Disjoint algorithm, we varied the size of the context feature vector ($d$) from $4$ to $32$, while keeping $N$ equal to $8$. For the hybrid algorithm, we varied the size of the action feature vector ($f$) from $4$ to $32$, keeping $d$ and $N$ equal to $8$.
    The following subsections describe the adopted methodology in detail and discuss the obtained results in terms of execution time and energy consumption.

    \subsection{Methodology}
    The standard software C++ implementation and the optimized version of both the Disjoint and Hybrid LinearUCB algorithms were compiled for the PULPino platform, the STM32 platform, and the Raspberry PI 4  platform. The compilation for PULPino Klessydra T13 was done for no vector computing support and for a $\text{SIMD}=4$ vector acceleration support. \conditionalhighlight{This SIMD width was selected as it offers a balance between performance and area, making it well-suited for embedded applications} \cite{Klessydra-T13}.
    The Klessydra PULPino platform was implemented on the Xilinx Vivado tool suite targeting the Kintex-7 KC705 FPGA device (xc7k325tffg900-2), with a target clock frequency of 100 MHz. The synthesis reported no timing violations.

    Execution time estimations were done by direct execution on the STM32 and Raspberry boards, while they were done by cycle-accurate RTL simulation on QuestaSim for the Klessydra T13 implementations, with 100 MHz simulated clock frequency. \conditionalhighlight{The speedup factor was calculated as the ratio between the execution times of the traditional algorithm and the optimized algorithm.}

    For energy consumption estimation, we measured the power absorbed by the STM32 and Raspberry Pi boards during algorithm execution using a power supply \conditionalhighlight{with integrated power measurement capabilities. The instrument, connected to an AC source and supplying the target boards with DC power, continuously monitored the absorbed current. After synchronizing with a digital trigger signal from the target boards, the average power consumption was recorded during algorithm execution.} For the Klessydra T13 implementations, we extracted the post-implementation netlist and conducted gate-level simulations of the algorithms. The simulations provided the switching activity \emph{.saif} files,  which were processed by the Vivado power estimator to assess the dynamic power $(P_{dynamic})$ of the core. For all of the Klessydra, Cortex A71 and Cortex M4 processors, the energy consumption was considered for the core and the memories only, excluding all the peripherals.
    For all the processors, we calculated the total energy consumption as the average power consumption multiplied by the execution time. \conditionalhighlight{The energy reduction factor was then calculated as the ratio between dynamic energy consumptions of the optimized algorithm and the traditional algorithm.}
        
    Tables \ref{tab: Disjoint_Table} and \ref{tab: Hybrid_Table} summarize all the obtained results for the Disjoint and Hybrid algorithms, on the chosen embedded execution platforms. The analysis of the specific impact of each optimization step is reported in the following.
    \begin{table}[!t]
        \centering
        \caption{Performance results for the Disjoint Algorithm in the Std. and Opt. versions on the examined execution platforms}
        \resizebox{\columnwidth}{!}{\begin{tabular}{|c|c|c|c|c|c|c|c|c|c|}
        \hline
        & \multirow{2}{*}{$d$} & \multicolumn{2}{c|}{Cortex M4 STM32 Nucleo} & \multicolumn{2}{c|}{Cortex A72 Raspberry PI 4} & \multicolumn{4}{c|}{Klessydra T13 PULPino} \\
        \cline{3-10}
        & & Std. & Opt. & Std. & Opt. & Std. & Opt. & Std. with VCU & Opt. with VCU \\
        \hline
        \multirow{4}{*}{\begin{tabular}[c]{@{}c@{}}Cycle\\ count\end{tabular}}
        & 4 & 433024 & 135586 & 198000 & 48000 & 96117 & 19078 & 27905 & 13371 \\
        & 8 & 2523904 & 467402 & 1117500 & 141000 & 484456 & 52767 & 131972 & 28854 \\
        & 16 & 19694862 & 1677126 & 7629000 & 487500 & 2734595 & 181357 & 498280 & 86287 \\
        & 32 & 148464048 & 6499592 & 55540500 & 1857000 & 17562142 & 682584 & 2017706 & 304447 \\
        \hline
        \multirow{4}{*}{\begin{tabular}[c]{@{}c@{}}Execution\\ time\\ $[\mu s]$\end{tabular}}
        & 4 	& 5412    & 1695  & 132       & 32 & 961 & 190 & 279 & 134 \\
        & 8 	& 31548   & 5843  & 745       & 94 & 4844 & 527 & 1320 & 289 \\
        & 16 	& 246185  & 20964 & 5086      & 325 & 27345 & 1813 & 4983 & 863 \\
        & 32 	& 1855800 & 81244 & 37027     & 1238 & 175621 & 6825 & 20177 & 3044 \\
        \hline
        \multirow{4}{*}{\begin{tabular}[c]{@{}c@{}}Energy\\ cons.\\ $[\mu J]$\end{tabular}}
        & 4  	& 838     & 262   & 435    & 105 & 185 & 40 & 54 & 27 \\
        & 8  	& 4890    & 905   & 2458   & 310 & 939 & 113 & 259 & 59 \\
        & 16 	& 38158   & 3249  & 16783  & 1072 & 5387 & 397 & 986 & 179 \\
        & 32 	& 287649  & 12592 & 122189 & 4085 & 35651 & 1549 & 4095 & 645 \\
        \hline
        \end{tabular}}
        \label{tab: Disjoint_Table}
    \end{table}

   \begin{table}[!t]
        \centering
        \caption{Performance results for the Hybrid Algorithm in the Std. and Opt. versions on the examined execution platforms}
        \resizebox{\columnwidth}{!}{%
        \begin{tabular}{|c|c|c|c|c|c|c|c|c|c|}
        \hline
        & \multirow{2}{*}{$f$} & \multicolumn{2}{c|}{Cortex M4 STM32 Nucleo} & \multicolumn{2}{c|}{Cortex A72 RASPberry PI 4} & \multicolumn{4}{c|}{Klessydra T13 PULPino} \\
        \cline{3-10}
        & & Std. & Opt. & Std. & Opt. & Std. & Opt. & Std. with VCU & Opt. with VCU \\
        \hline
        \multirow{4}{*}{\begin{tabular}[c]{@{}c@{}}Cycle\\ count\end{tabular}}
        & 4  & 5751676 & 4355380 & 2325000 & 1785000 & 1029600 & 909378 & 413274 & 327463 \\
        & 8  & 28419702 & 14498906 & 11142000 & 5679000 & 4594114 & 2919300 & 1307512 & 974813 \\
        & 16 & 176841698 & 52908188 & 67779000 & 20509500 & 26216770 & 10454626 & 4741130 & 3338271 \\
        & 32 & N.A. & N.A. & 456472500 & 78625500 & 176168851 & 39638697 & 19156387 & 12768900 \\
        \hline
        \multirow{4}{*}{\begin{tabular}[c]{@{}c@{}}Execution\\ time\\ $[\mu s]$\end{tabular}}
        & 4  	& 71895 		& 54442 	& 1550 	& 1190 	& 10296 	& 9093 & 4133 & 3275 \\
        & 8  	& 355246 		& 181236 & 7428 	& 3786 	& 45941 	& 29193 & 13075 & 9748 \\
        & 16 	& 2210521 		& 661352 & 45186 	& 13673 	& 262167 & 104546 & 47411 & 33383 \\
        & 32 	& N.A. 			& N.A. 		& 304315 & 52417 	& 1761688& 396487 & 191564 & 127689 \\
        \hline
        \multirow{4}{*}{\begin{tabular}[c]{@{}c@{}}Energy\\ cons.\\ $[\mu J]$\end{tabular}}
        & 4 	& 11143 		& 8438 		& 5115 		& 3927 		& 1945 		& 1746 		& 847 		& 677 \\
        & 8 	& 55063 		& 28091 	& 24512 	& 12493 	& 8912 		& 5634 		& 2706 		& 2037 \\
        & 16 	& 342630 		& 102509 	& 149113 	& 45120 	& 53482 	& 20700 	& 10193 	& 7344 \\
        & 32 	& N.A. 			& N.A. 		& 1004239 	& 172976 	& 389333 	& 80466 	& 43676 	& 30006 \\
        \hline
        \end{tabular}%
        }
        \label{tab: Hybrid_Table}
    \end{table}
    
    \subsection{Algorithmic optimization impact on execution time}

\begin{figure}[!t]
            \centering
            \begin{subfigure}[b]{0.48\columnwidth}
                \Description{The proposed optimization on the Disjoint algorithm achieves a speedup that ranges from $5.04\times$ to $25.73\times$ on Klessydra T13, from $3.19\times$ to $22.84\times$ on Cortex M4 and from $4.13\times$ to $29.91\times$ on Cortex A72, when varying the context feature vector size from 4 to 32.}
                \includegraphics[width=\textwidth]{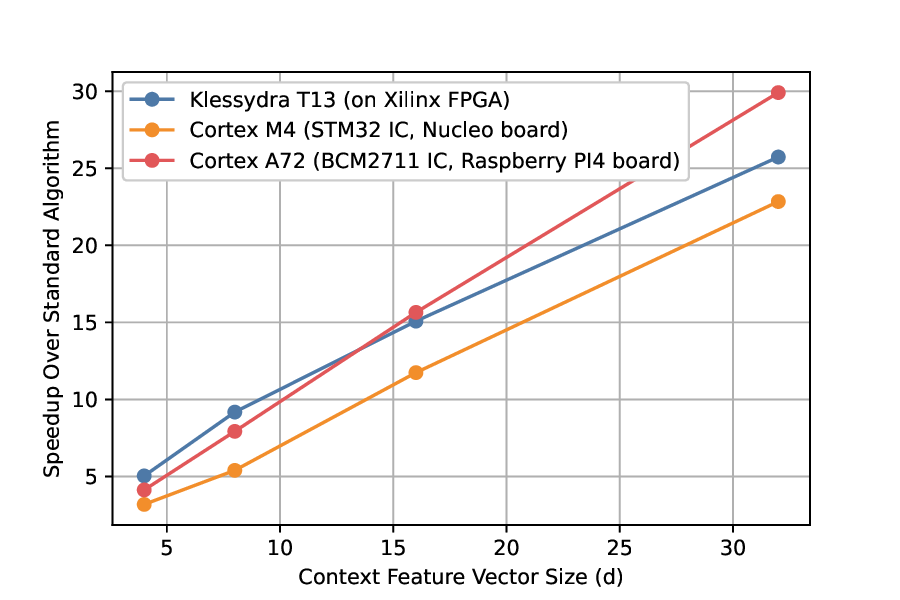}
                \caption{Speedup for the Disjoint algorithm}
                \label{fig:alg_opt_disj_time}
            \end{subfigure}
            \hspace{0.02\columnwidth} 
            \begin{subfigure}[b]{0.48\columnwidth}
                \Description{In the Hybrid algorithm, the algorithmic optimization based on the Sherman-Morrison-Woodbury formulas results in speedups ranging from $1.13 \times$ to $4.45 \times$ on the Klessydra-T13 core, and from $1.30 \times$ to $5.80 \times$ on the Cortex-A7. For the Cortex-M4 core, the speedup increases from $1.32 \times$ at $f=4$ to $3.34 \times$ at $f=16$. Notably, data for $f=32$ are absent due to the high memory demands of this configuration, which exceed the capabilities of the target Nucleo Board.}
                \includegraphics[width=\textwidth]{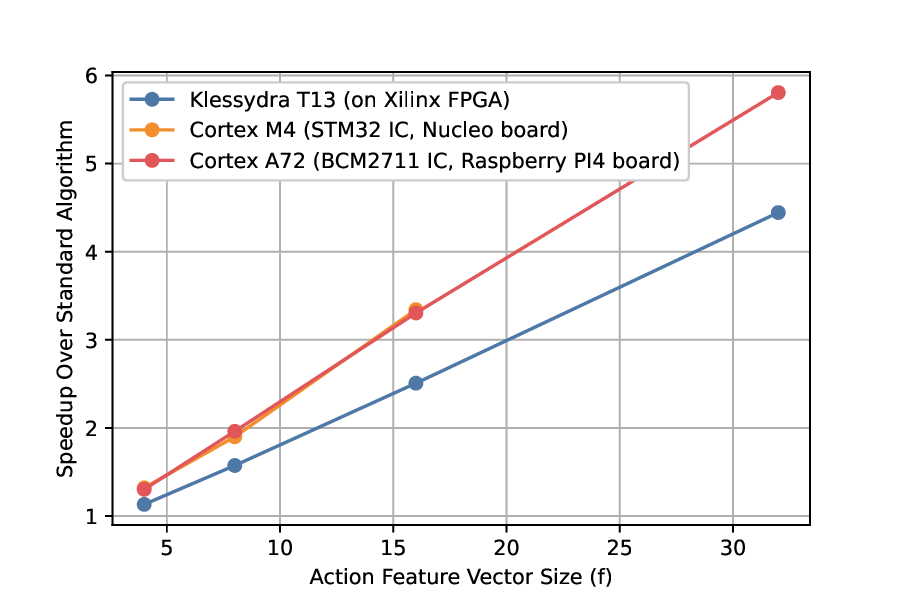}
                \caption{Speedup for the Hybrid algorithm}
                \label{fig:alg_opt_hyb_time}
            \end{subfigure}
            \caption{Algorithmic optimization impact on execution time}
            \label{fig:combined_alg_speedups}
        \end{figure}

        Efficient execution time is essential for deploying AI algorithms in applications with stringent real-time requirements.
        Figure \ref{fig:alg_opt_disj_time} illustrates the impact of the Sherman-Morrison algorithmic optimization on the Disjoint LinearUCB algorithm with $N$ equal to 8. Varying the context feature vector size ($d$) from $4$ to $32$ the proposed approach achieves a speed up that ranges from $5.04\times$ to $25.73\times$ on Klessydra T13, from $3.19\times$ to $22.84\times$ on Cortex M4 and from $4.13\times$ to $29.91\times$ on Cortex A72. 
        \conditionalhighlight{The observed linear increase in speedup with $d$
        aligns with the theoretical reduction in complexity from $\mathcal{O}(d^3)$ to $\mathcal{O}(d^2)$ described in Table 1 and Fig.} \ref{fig: Disjoint_complexity}.
        These results highlights how the removed matrix inversions constitute the performance bottleneck of the algorithm and strongly limits its scalability. Notably, as described in Section 3, increasing the number of actions, $N$, would further enhance the speedup provided by the algorithmic optimization.
        
        The same analysis on the execution time is depicted in Figure \ref{fig:alg_opt_hyb_time} for the Hybrid algorithm, varying the action feature vector size ($f$) from $4$ to $32$, while keeping $d$ and $N$ equal to $8$. The algorithmic optimization based on the Sherman-Morrison-Woodbury formulas results in speedups ranging from $1.13 \times$ to $4.45 \times$ on the Klessydra-T13 core, and from $1.30 \times$ to $5.80 \times$ on the Cortex-A7. For the Cortex-M4 core, the speedup increases from $1.32 \times$ at $f=4$ to $3.34 \times$ at $f=16$. Notably, data for $f=32$ are absent due to the memory demands of this configuration, which exceed the capabilities of the target Nucleo Board.  \conditionalhighlight{The results demonstrate a linear increase in speedup with $f$, consistent with the theoretical reduction in complexity from $\mathcal{O}(f^3d^3)$ to $\mathcal{O}(f^2d^3)$, as outlined in Table 1 and Fig.} \ref{fig: Hybrid_complexity}.
        As detailed in Figure \ref{fig: Hybrid_complexity}, increasing the value of $d$ would increase the speedup for the optimized Hybrid algorithm, whereas increasing $N$ has a minimal impact on performance due to the dominance of arithmetic operations on the complex $k\times k$ shared matrix.

        \begin{figure*}[!t]
            \centering
            \begin{subfigure}[b]{0.48\textwidth}
                \Description{For the optimized Disjoint algorithm, on the Klessydra-T13 core, the energy reduction factors range from $0.22$ to $0.04$ when d is equal to 4 and 32, respectively.  A similar pattern in energy efficiency can be observed for the Cortex-M4 and Cortex-M7 platforms, with energy reduction factors ranging from $0.31$ to $0.04$ and from $0.24$ to $0.03$, respectively.}
                \includegraphics[width=\textwidth]{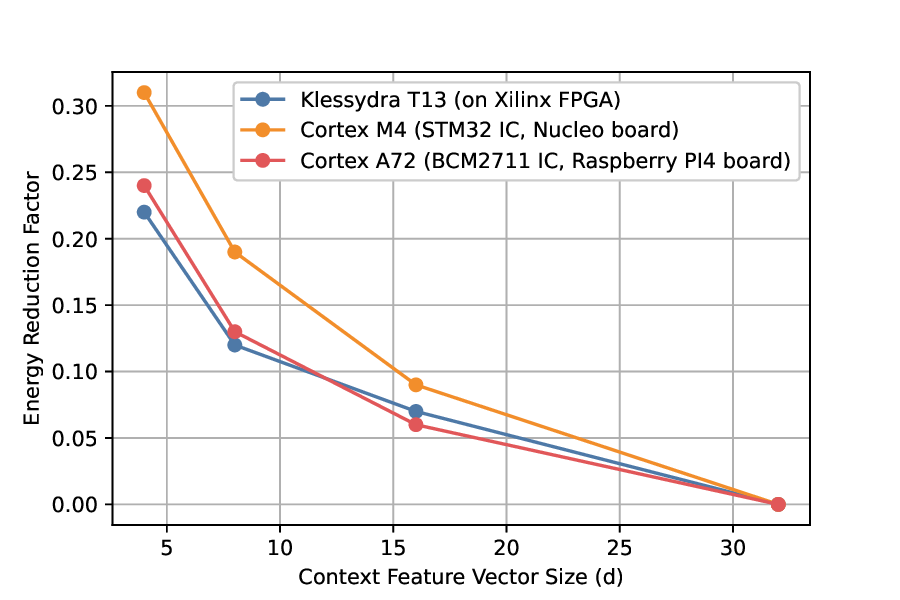}
                \caption{Energy reduction factor for the Disjoint algorithm}
                \label{fig:power_disj}
            \end{subfigure}
            \hspace{0.02\textwidth} 
            \begin{subfigure}[b]{0.46\textwidth}
            \Description{In the Hybrid algorithm, varying the action feature vector size ($f$) from $4$ to $32$, while keeping $d$ and $N$ equal to $8$, the energy reduction decreases from $0.89$ to $0.20$ on Klessydra. As already discussed for the execution time, the data for $f=32$ with Cortex-M4 are not present due to the high memory demands of this configuration, but the trend for this platform closely follows the one obtained for the Cortex-A72, in which the energy reduction factor ranges from $0.76$ to $0.17$. }
                \includegraphics[width=\textwidth]{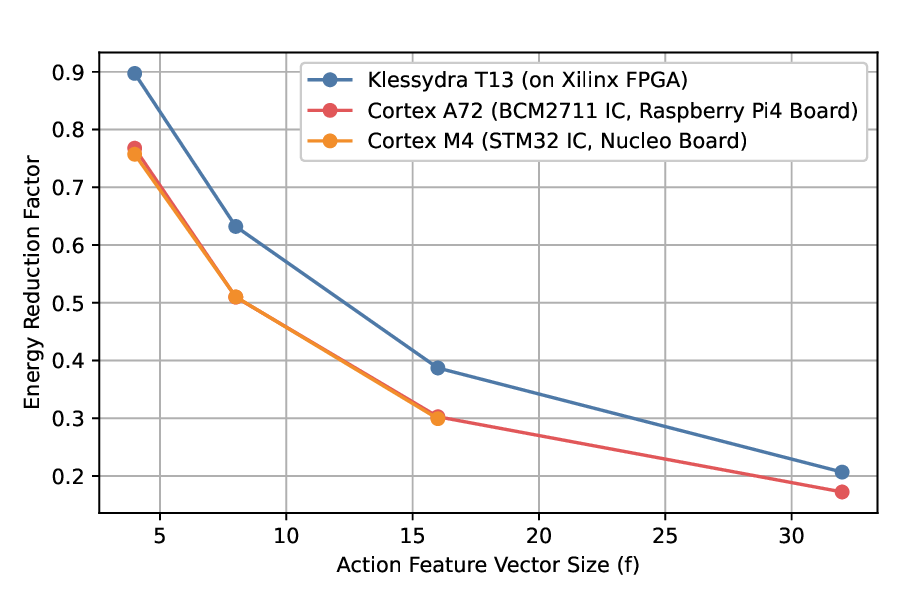}
                \caption{Energy reduction factor for the Hybrid algorithm}
                \label{fig:power_hyb}
            \end{subfigure}
            \caption{Algorithmic optimization impact on energy consumption}
            \label{fig:combined_alg_energy}
        \end{figure*}

    \subsection{Algorithmic optimization impact on energy consumption}\label{Power_consumption}
        Energy consumption is another critical factor when deploying AI algorithms in low-power environments. This section examines how the proposed algorithmic optimizations enhance the energy efficiency of the LinearUCB algorithms.

        Figure \ref{fig:power_disj} depicts the results obtained for the Disjoint algorithm varying the context feature vector size ($d$) while keeping $N$ equal to $8$. As already observed for the execution time, as $d$ increases, so does the advantage brought by algorithmic optimizations.
        \conditionalhighlight{The energy reduction factor decreases asymptotically as the ratio  $\frac{\mathcal{O}(d^2)}{\mathcal{O}(d^3)} \propto \frac{1}{d}$, aligning with the theoretical analysis of the computational complexity presented in Table 1 and Fig.}\ref{fig: Disjoint_complexity}.       
        On the Klessydra-T13 core, the energy reduction factors ranges from $0.22$, to $0.04$ when d is equal to 4 and 32, respectively.  A similar pattern in energy efficiency can be observed for the Cortex-M4 and Cortex-M7 platforms, with energy reduction factors ranging from $0.31$ to $0.04$ and from $0.24$ to $0.03$, respectively.

        For the Hybrid algorithm the energy reduction obtained by the algorithmic optimization is shown in Figure \ref{fig:power_hyb} varying the action feature vector size ($f$) from $4$ to $32$, while keeping $d$ and $N$ equal to $8$. 
        \conditionalhighlight{The trend shows an asymptotic decrease as the ratio $\frac{\mathcal{O}(f^2d^3)}{\mathcal{O}(f^3d^3)} \propto \frac{1}{f}$, as outlined in Table 1 and Fig.} \ref{fig: Hybrid_complexity}. 
        On Klessydra, the energy reduction decreases from $0.89$ to $0.20$. As already discussed for the execution time, the data for $f=32$ with Cortex-M4 are not present due to the high memory demands of this configuration, but the trend for this platform closely follows the one obtained for the Cortex-A72, in which the energy reduction factor ranges from $0.76$ to $0.17$. 

        \begin{figure*}[!t]
            \centering
            \begin{subfigure}[b]{0.48\textwidth}
            \Description{ In the Disjoint version, vector computing acceleration achieves speedups ranging from \(3.44\times\) at \(d=4\) to \(8.70\times\) at \(d=32\) for the standard version. When combined with algorithmic optimizations, these speedups escalated dramatically, ranging from \(7.18\times\) to \(57.68\times\).}
                \includegraphics[width=\textwidth]{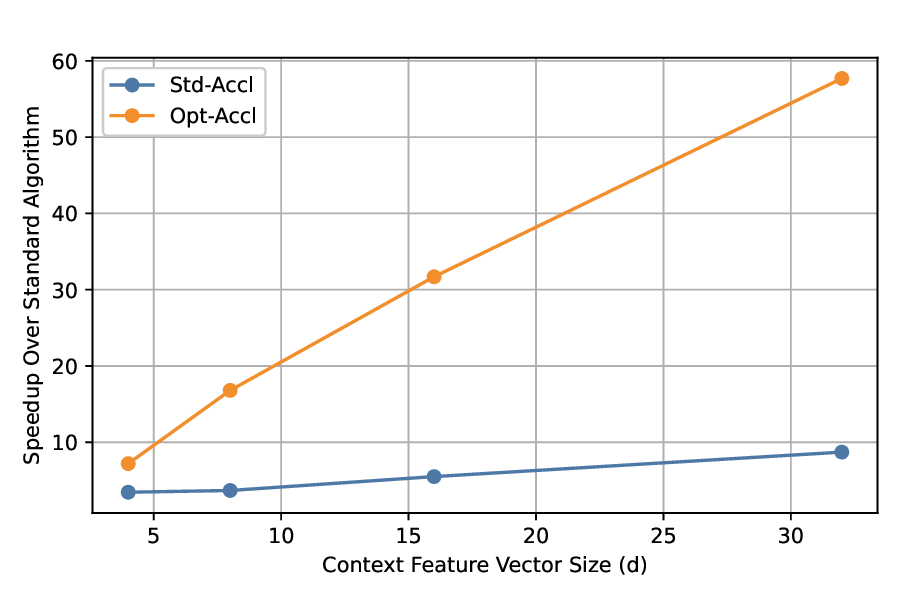}
                \caption{Speedup for the Disjoint algorithm}
                \label{fig:accl_opt_disj_time}
            \end{subfigure}
            \hspace{0.02\textwidth} 
            \begin{subfigure}[b]{0.48\textwidth}
                \Description{For the Hybrid algorithm, the vector computing acceleration applied to the standard version of the algorithm results in speedups ranging from $2.49 \times$ when $f= 4$ to $9.19 \times$ at $f=32$. When vector acceleration and algorithmic optimization are synergically applied, the speedup ranges from $3.15\times$ to $13.82\times$}
                \includegraphics[width=\textwidth]{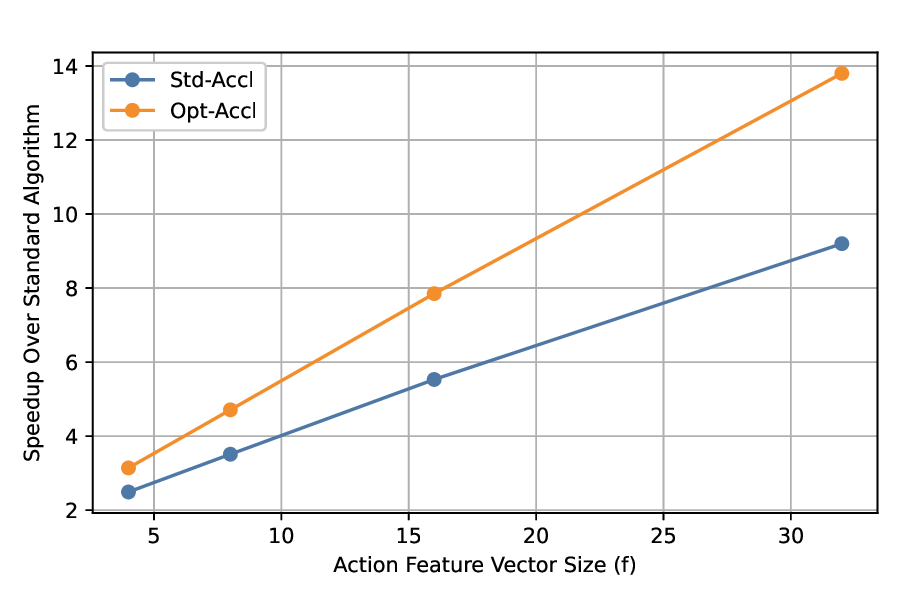}
                \caption{Speedup for the Hybrid algorithm}
                \label{fig:accl_opt_hyb_time}
            \end{subfigure}
            \caption{Vector computing acceleration impact on execution time}
            \label{fig:combined_vec_speedups}
        \end{figure*}
        
        \begin{figure*}[!t]
            \centering
            \begin{subfigure}[b]{0.48\textwidth}
            \Description{For the standard Disjoint algorithm, the energy reduction factor decreases from $0.29$ to $0.11$ as \(d\) is increased from 4 to 32. The optimized version further reduces these values from $0.15$ to $0.02$.}
                \includegraphics[width=\textwidth]{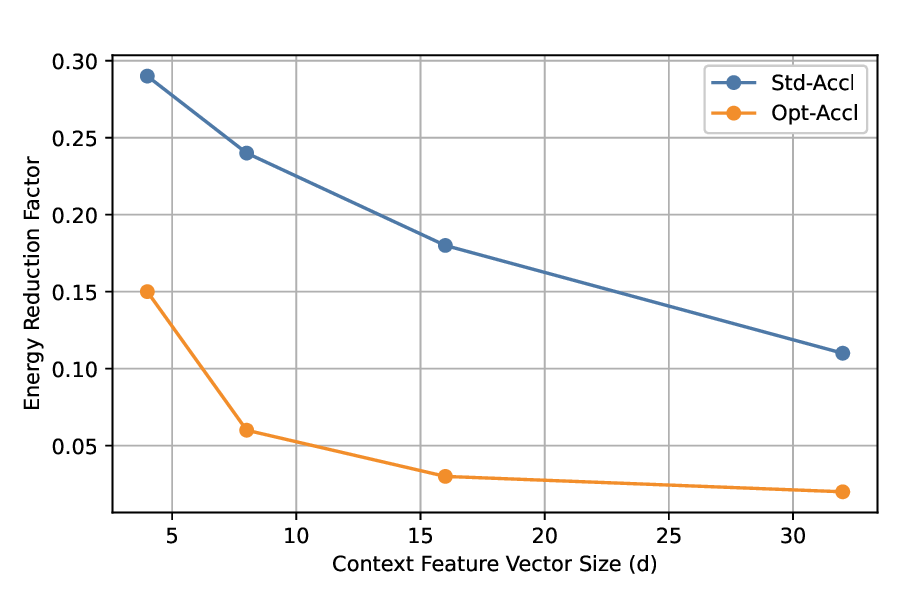}
                \caption{Energy reduction factor for the Disjoint algorithm}
                \label{fig:power_disj_accl}
            \end{subfigure}
            \hspace{0.02\textwidth} 
            \begin{subfigure}[b]{0.48\textwidth}
                \Description{The effect of vector acceleration is even more pronounced in the Hybrid algorithm, where the energy reduction factor ranges from $0.44$ to $0.11$ for the standard version and from $0.35$ to $0.08$ for the optimized one. }
                \includegraphics[width=\textwidth]{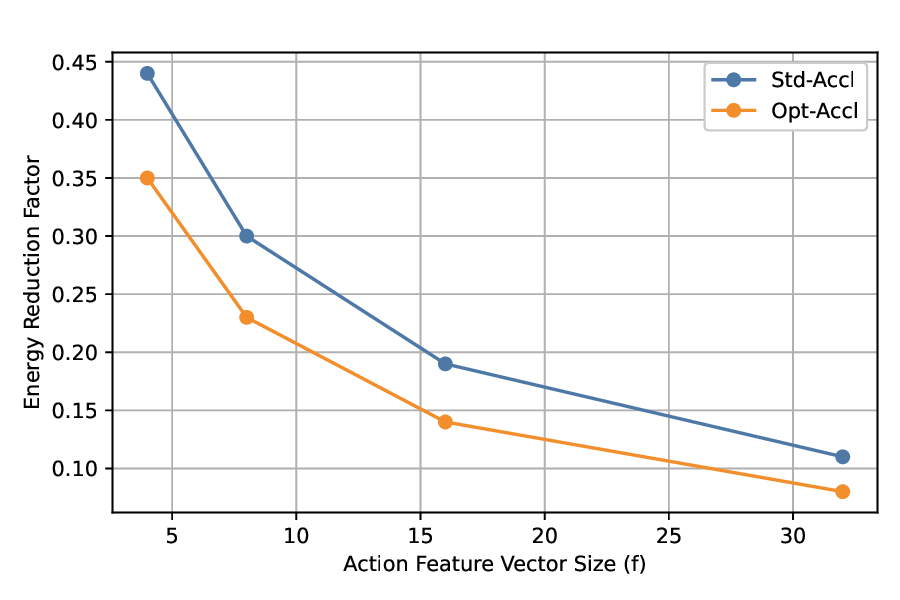}
                \caption{Energy reduction factor for the Hybrid algorithm}
                \label{fig:power_hyb_accl}
            \end{subfigure}
            \caption{Vector computing acceleration impact on energy consumption}
            \label{fig:combined_vec_energy}
        \end{figure*}

    \subsection{Vector computing acceleration impact on execution time}
        The impact of vector computing support on the LinearUCB algorithms was evaluated by executing both the standard and optimized versions on the Klessydra-T13 core, utilizing the VCU support for accelerating the basic matrix operations, as detailed in Section \ref{Hardware Acceleration}. The execution times have been measured and compared with the ones obtained by software executions and the results in terms of speedup are reported in Figures \ref{fig:accl_opt_disj_time} and \ref{fig:accl_opt_hyb_time} for the Disjoint and Hybrid algorithms, respectively.

        In the Disjoint version, vector computing acceleration achieves speedups ranging from \(3.44\times\) at \(d=4\) to \(8.70\times\) at \(d=32\) for the standard version. When combined with algorithmic optimizations, these speedups escalated dramatically ranging from \(7.18\times\) to \(57.68\times\).
    
        For the Hybrid algorithm, the vector computing acceleration applied to the standard version of the algorithm results in speedups ranging from $2.49 \times$ when $f= 4$ to $9.19 \times$ at $f=32$. When vector acceleration and algorithmic optimization are synergically applied, the speedup ranges from $3.15\times$ to $13.82\times$, resulting in essential improvement for deploying complex algorithms like the Hybrid one on embedded systems.

        \conditionalhighlight{Notably, the observed trends for the accelerated standard versions confirm the expected linear speedup provided by vector support when varying the matrix size observed in Fig.}\ref{fig: Matrix_operations}\conditionalhighlight{. Since both the algorithmic optimization and the vector acceleration independently contribute linear scaling properties, their combination results in an overall effect on speedup that remains linear but with a greater magnitude.}

    \subsection{Vector computing acceleration impact on energy consumption}
        Finally, the impact of vector computing support on the energy consumption of the target algorithms is depicted in Figures \ref{fig:power_disj_accl} and \ref{fig:power_hyb_accl}.
        For the standard Disjoint algorithm, the energy reduction factor decreases from $0.29$ to $0.11$ as \(d\) is increased from 4 to 32. The optimized version further reduces these values from $0.15$ to $0.02$.
        
        The effect of vector acceleration is even more pronounced in the Hybrid algorithm, where the energy reduction factor ranges from $0.44$ to $0.11$ for the standard version and from $0.35$ to $0.08$ for the optimized one. 
        
        \conditionalhighlight{The observed reductions align with theoretical expectations of an inverse relationship with $d$ and $f$ due to the linear scaling properties of vector acceleration when varying matrix sizes, as demonstrated in Fig.} \ref{fig: Matrix_operations}.

    \subsection{Discussion}    
        Overall, the obtained results demonstrate the effectiveness of the dual-pronged optimization strategy, which combines algorithmic improvements with vector processing support to reduce execution time and energy consumption, enabling faster and more power efficient application of LinearUCB algorithms in embedded learning systems.
        For the Disjoint algorithm, the algorithmic refinements based on the Sherman-Morrison formula replace heavy matrix inversions with lightweight incremental updates, resulting in significant execution time improvements across all platforms, with a maximum speedup on Klessydra equal to $25.73\times$. At the same time, although less effective in this case, vector computing support achieves a maximum speedup of $8.70\times$. Combined, the total speedup reaches $58\times$.
        A different trend can be observed for the Hybrid algorithm, which involves larger shared matrices and is inherently more complex. In this case, the Sherman-Morrison-Woodbury-based algorithmic optimization provides a maximum speedup of $4.45\times$, and the vector computing acceleration achieves $9.19\times$. Together, they yield a total speedup of $13.82\times$.
        Overall, increasing the size of context and action feature vectors enhances the advantage of vector parallel processing. The presented dual-pronged strategy effectively improves execution time and energy consumption, proving beneficial for deploying AI models in edge computing environments with low-power and real-time requirements.

%% file: Sections/6.Conclusions.tex
    \section{Conclusions}\label{Conclusions}
    In this work, we introduced a dual optimization strategy that combines algorithmic and hardware support techniques to cut down on the execution time and energy consumption of two LinearUCB Contextual Bandits algorithms, in the context of embedded learning systems.
    Using the Sherman-Morrison-Woodbury formula, we replaced intensive matrix inversions with efficient incremental updates, reducing memory requirements and computational complexity.
    Porting the code on the Klessydra-T13 RISC-V core's vector computing support, we expedited core matrix operations, improving speed and energy efficiency.
    To validate the proposed enhancements, we implemented and executed both traditional and optimized LinearUCB algorithms on the Cortex M4 STM32 platform, the Cortex A72 Raspberry Pi 4 Model B platform, and the RISC-V Klessydra-T13 PULPino platform, varying context and action feature vector sizes to evaluate execution speedup and energy reduction.
    The obtained results demonstrate substantial improvements, making this integrated approach highly suitable for embedded applications with stringent execution time and power consumption requirements, like learning systems on the edge of the Internet-of-Things.